\documentclass{article}
\usepackage{iclr2027_conference,times}

\usepackage[utf8]{inputenc}
\usepackage[T1]{fontenc}
\usepackage[hidelinks]{hyperref}
\usepackage{url}
\usepackage{booktabs}
\usepackage{array}
\newcolumntype{R}[1]{>{\raggedright\arraybackslash}p{#1}}
\usepackage{amsfonts}
\usepackage{amssymb}
\usepackage{xcolor}
\usepackage{amsmath}
\usepackage{nicefrac}
\usepackage{microtype}
\usepackage{graphicx}
\usepackage{multirow}
\usepackage{float}
\usepackage{rotating}
\usepackage[labelfont=bf]{caption}
\usepackage{tikz}
\usetikzlibrary{arrows.meta,positioning,fit,calc}

\graphicspath{{figs/}}

\title{Which Decisions Low-Bit Quantization Breaks,\\and How to Predict Them}

\author{Zekun Wu$^{1,2}$ \quad Swati Dhiman$^{1,2}$ \quad Adriano Koshiyama$^{1}$ \\
$^{1}$Holistic AI \quad $^{2}$UCL Centre for Artificial Intelligence, University College London \\
\texttt{\{zekun.wu, swati.dhiman, adriano.koshiyama\}@holisticai.com}}

\iclrfinalcopy

\begin{document}

\maketitle
\lhead{Preprint. Under review.}

\begin{abstract}
Quantization saves memory by storing model weights with fewer bits. It can also change model decisions, such as whether to call a tool or which option to choose from a finite set. We study these decision changes in 16 language models from 8 families at 4, 3 and 2 bits, across several post-training quantization settings. Our evaluation covers tool use, safety, general knowledge and social bias, using BFCL, XSTest, MMLU, BoolQ, BBQ and synthetic tasks. The decision \emph{margin} is the score difference between two possible first tokens, measured before and after quantization. Its size shows how strongly the model prefers one option, and a change in sign means the preference has flipped. Writing the margin before quantization as $m$ and the margin after quantization as $m'$, we find an approximately linear relationship across decisions: $m' \approx c\,m + b$. The slope $c$ is usually below one and becomes smaller as precision falls, so quantization progressively shrinks decision margins. The offset $b$ is the same for every decision of one kind. Quantization therefore does not simply add random noise, and even a strong preference at full precision can flip. Quantization also affects different kinds of decisions to different degrees. Within tool use, whether to call a tool is often more sensitive than which tool to call: on 400 BFCL tasks, three of five models lose more completed calls than correct tool selections at 3-bit round-to-nearest. Under GPTQ and GGUF far fewer whether-to-call decisions flip than under plain rounding, so there is no single 3-bit failure point. The same relationship predicts how often decisions flip. Across 1{,}082 combinations of models, quantization settings, bit-widths and decision types drawn from our evaluation, we fit the slope, the offset and the spread around the fitted line on half of the decisions and predict the flip rate on the other half. The predicted flip rate differs from the observed flip rate by a median of 1.0 percentage point, while reusing the flip rate of the first half misses by 1.3. Together, these results show a common pattern in how quantization changes decisions and provide a way to predict when those changes will occur.
\end{abstract}

\section{Introduction}

Post-training quantization reduces the memory required to run large language models \citep{dettmers2023kbit,frantar2022gptq}, but it can also change the decisions they make. For tool-using agents, whether to call a tool and which tool to call are distinct decisions, and quantization need not affect them equally. After 3-bit rounding, Qwen3-4B still prefers the correct tool on the requests we wrote, yet of the 77 requests that need a tool and that it answered with a tool call at full precision, it now calls a tool on none (Appendix~\ref{sec:instrument}). We investigate how quantization changes such preferences and whether these changes can be predicted.

\begin{figure}[t]
\begin{center}
\includegraphics[width=\linewidth]{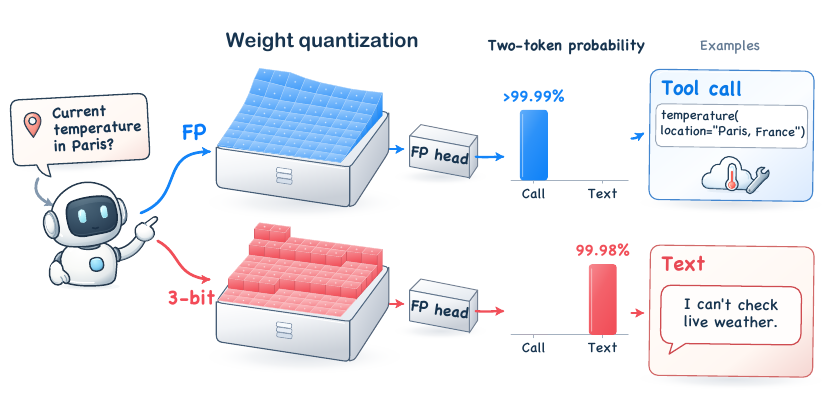}
\end{center}
\caption{\textbf{A tool-call preference reverses after 3-bit quantization.} For the same weather request, Qwen3-4B favors the tool-call opening token at full precision (blue) and the text-opening token \texttt{I} after 3-bit rounding (red). The surfaces show sorted magnitudes of the same weight group before and after rounding. The output head stays at full precision in both. Bars show probabilities normalized over these two tokens only. Responses on the right are illustrative examples.}
\label{fig:overview}
\end{figure}

Quantization methods are designed around weight and layer-output error: \citet{nagel2020adaround} and \citet{li2021brecq} choose rounding values by their effect on a layer's output, and \citet{lin2023awq} rescales salient weight channels. Neither error says which decisions will change.

We study 16 language models from 8 families at 4, 3 and 2 bits using round-to-nearest. We also evaluate three other quantization methods on all 16 models, and widths up to 8 bits on subsets of these models. Our evaluation covers tool use, safety refusal, general knowledge and social bias. Within tool use, we examine four choices separately: whether to call a tool, which tool to call, how to fill its arguments, and how to use its returned result.

We study each \emph{decision} as a choice between two options represented by distinct first tokens. Their \emph{margin} is the difference between the model's scores (logits) for those tokens. A \emph{flip} occurs when quantization reverses which token has the higher score. The \emph{flip rate} is the share of items that flip, counting only items whose margin at full precision is not zero. Figure~\ref{fig:overview} illustrates one such reversal.

We find that quantization does more than add random noise to decision margins. Within one kind of decision, the quantized margin is approximately a linear function of the original: shrunk by a factor that falls as precision falls, and shifted toward one option. The damage is also uneven across kinds of decision: at 3 bits under round-to-nearest, whether to call a tool collapses on most models, while which tool to call changes less.

In \S\ref{sec:forecast}, we test whether this relationship predicts how often preferences reverse on new items. For each model, quantization configuration and kind of decision, we fit the relationship on half of the items. We then predict the fraction of the remaining items whose preferences will reverse. The forecast has a lower median error than using the fraction that reversed in the first half.

\paragraph{Contributions.}
(i)~An evaluation across tool use, safety refusal, general knowledge and social bias. The evaluation shows that quantization disrupts whether to call a tool more than which tool to choose.
(ii)~An approximately linear relationship between margins before and after quantization: quantization often shrinks the original margin and shifts it toward one option. We test this relationship against three additive-noise models.
(iii)~A forecast of preference reversals, fitted on half of the items and evaluated on the other half. It beats the first half's observed flip rate.

\section{Measurement and experimental design}
\label{sec:measure}
\subsection{Decision-level measurement}
\label{sec:decision-measurement}
For the weather request in Figure~\ref{fig:overview}, the model can open a tool call or start a text reply. We score the first token of each option on the same prompt, first at full precision and then after quantization.

\begin{samepage}
Each \emph{item} is a prompt $x$ with two candidate tokens, $t_1$ and $t_2$. Their full-precision margin is
\begin{equation}
m(x)=\ell_{\mathrm{FP}}(t_1\mid x)-\ell_{\mathrm{FP}}(t_2\mid x),
\label{eq:margin}
\end{equation}
where $\ell_{\mathrm{FP}}$ is the token's score (logit). A positive margin favors $t_1$ and a negative one $t_2$. We compute the same difference after quantization, denoted by $m'(x)$. Each item thus gives one pair $(m,m')$. Opposite signs indicate a flip. Fits and flip rates use only items whose full-precision margin is not zero, which leaves out items whose two options start with the same token.
\end{samepage}

For most tasks, $t_1$ begins the correct answer and $t_2$ begins the highest-scoring alternative at full precision. We keep this pair fixed after quantization. Whether-to-call instead compares calling with replying in text, and the two refusal tasks compare declining with complying. In these tasks, either option can be correct, and we record the correct option as the item's \emph{side}. A flip can therefore add an error or fix one.

A margin measures preference between two tokens. Appendix~\ref{sec:instrument} checks agreement with generated tool calls and refusals. This correspondence weakens under severe quantization.

\subsection{Experimental design}
\label{sec:experimental-design}
\noindent\textbf{Benchmarks and decisions.} Four test sets hold 1{,}092 items from benchmark subsets and authored tasks, covering eleven kinds of decision. Tool calling is the main case: BFCL \citep{bfcl2024} supplies whether-to-call requests, while authored tasks isolate tool selection, argument filling and tool-result use. A separate 400-task BFCL evaluation measures call completion and tool selection on the same requests. Its five models pass the full-precision capability checks specified in Appendix~\ref{app:crossed}. Public which-tool tasks from BFCL and MetaTool \citep{huang2024metatool} and argument-fill tasks from BFCL measure the same two decisions on public data (Table~\ref{tab:decisions}). To examine decisions beyond tool use, we use XSTest for refusal \citep{rottger2024xstest}, MMLU and BoolQ for knowledge and question answering \citep{hendrycks2021mmlu,clark2019boolq}, and BBQ for social bias \citep{parrish2022bbq}. Synthetic arithmetic and code tasks add procedural choices, including \emph{code tokens}, the choice of the next symbol in a short piece of code. Appendix~\ref{app:setupfull} gives sampling, counts and token pairs.

\noindent\textbf{Models.} The weight experiments cover 16 models from eight families: Qwen2.5, Qwen3, Qwen3.5, Granite, Gemma-3, Gemma-4, Llama-derived models and Phi. Model sizes range from 0.6B to 32B. Four additional base-model controls test whether the pattern also occurs before instruction tuning.

\noindent\textbf{Quantization and controls.} Weight quantization targets decoder-block linear layers. Embeddings and output heads remain at full precision. We evaluate:
\begin{itemize}
\item \textbf{Round-to-nearest (RTN):} 4, 3 and 2 bits on all 16 models, with 8-, 6- and 5-bit sweeps on five models.
\item \textbf{Activation-aware scaling:} 4, 3 and 2 bits on all 16 models.
\item \textbf{GPTQ:} 4, 3 and 2 bits on all 16 models.
\item \textbf{GGUF:} nominal 8-, 4-, 3- and 2-bit versions of all 16 models.
\end{itemize}
Additional experiments vary activation and KV-cache precision, test rotation and repairs, and check alternative token pairs and margin-matched items. Vision--language tasks and a set of agent-oversight questions test the measurement's validity in additional domains (Appendix~\ref{app:edges}). Appendix~\ref{app:setuplist} gives the full quantization coverage.

\noindent\textbf{Analysis protocol.} A \emph{condition} is one model configuration, bit-width and kind of decision in one run. Calling and refusal are also analyzed by side. Appendix~\ref{app:accounting} details the counting. We fit on half of the items and predict flips on the other half, also checking a split that keeps repeated requests together. Section~\ref{sec:forecast} reports the forecast, baseline and cross-model transfer comparisons.

\section{The damage is a fraction, not an amount}

\label{sec:attenuation}

\paragraph{What the plot shows.}

\begin{figure}[t]
\begin{center}
\includegraphics[width=\linewidth]{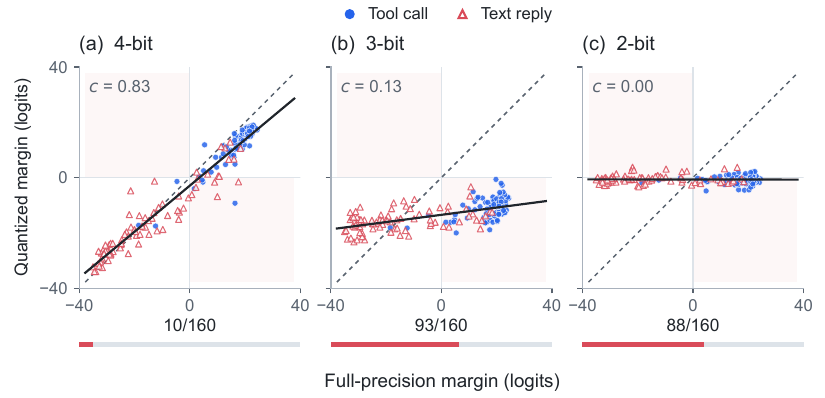}
\end{center}
\caption{Quantized against full-precision margins (Qwen3-4B, whether-to-call,
round-to-nearest). All three panels use identical square axes. Dashed lines are the identity,
solid lines are fitted responses, and pale quadrants indicate sign changes. Colour gives each item's correct response, its should-call or should-not-call side. The counts and strips below report flips among the same 80 items a side. Labels give fitted slopes $c$. Against the additive account, the cloud tips toward the horizontal instead of keeping
the identity's angle.}
\label{fig:scatter}
\end{figure}

Figure~\ref{fig:scatter} puts quantized margins against full-precision ones. The relation passes through three stages. At 4 bits the points sit on a line whose slope is a little below one: everything has shrunk by about a sixth and shifted down by three logits, and almost nothing has changed sides. At 3 bits the line is nearly flat and most of what was positive is now negative. The two sides separate into clusters there. The separate fits in Appendix~\ref{app:sideflips} give low slopes on both sides, so the pooled slope is shrinkage within each side and not the gap between them. At 2 bits the line is flat: the fitted slope is zero.

Additive noise would give a band of constant width along the diagonal. The data give a line with a slope below one and an intercept away from zero. We write the fitted relation as
\begin{equation}
m' = c\cdot m + b + \varepsilon, \qquad \varepsilon \sim \mathcal{N}(0, \sigma^2).
\label{eq:law}
\end{equation}
$c$ is the \emph{surviving fraction}: 1 when nothing is lost, 0 when the quantized margin no longer depends on the original. Below zero, the quantized margin points the opposite way to the original. One condition in six where $c$ can be estimated has it below zero. In the main matrix, mostly round-to-nearest and activation-aware runs (Appendix Table~\ref{tab:accounting}), its median is 0.86 at 4 bits, 0.29 at 3 and 0.00 at 2. The \emph{offset} $b$ moves every margin in the same direction. On tool calls the offset is a penalty on the one token that opens the call, which \S\ref{sec:dissociation} shows by scoring the whole vocabulary on Qwen3-4B. A drift shared by every logit would cancel in a margin. $\sigma$ is the residual spread. All three are fitted per condition. We call Eq.~\ref{eq:law} \emph{the law}, and test it below against three additive alternatives.  Flips come from all three terms: shrinkage removes the protection a large margin gave, the offset decides which way the decision falls, and the residual flips the smallest margins either way.

Appendix Figure~\ref{fig:spread} shows that the surviving fraction falls steeply between 4 bits and 2 in all 14 models plotted. The model is not simply turning its volume down: rescaling every logit by one factor would change no decision. The damage is also uneven: on Qwen3-4B under 3-bit round-to-nearest, the surviving fraction is 0.94 for code tokens against 0.13 for whether-to-call (Appendix Table~\ref{tab:decisionsfull} gives each kind's median over models).

\paragraph{Ruling out the additive story.}
\label{sec:rivals}

An additive model whose noise grows with the margin draws the same picture as a slope below one. So we fit four competing accounts by maximum likelihood and let BIC pick between them. BIC rewards a better fit and charges for every extra parameter. Two of them are constant-size noise, with and without a drift term. The third is additive noise whose size grows with the margin, which we call \emph{margin-scaled noise}. The fourth is the law itself, the multiplicative account.

A condition is \emph{damaged} when its fitted $c$ is below 0.70. That definition is about how much of the margin is left, not about whether any decision changed sides: a kind of decision whose margins are wide can lose most of them and still flip nothing. The multiplicative account wins all 642 damaged conditions. This comparison extends beyond the four main test sets to include the agent-oversight questions and vision--language tasks, the external tasks of \S\ref{sec:dissociation} and its margin-matched controls. Appendix Table~\ref{tab:accounting} lists the data behind the main analyses. The cutoff sensitivity checks in Appendix~\ref{app:rivals} leave the multiplicative account the winner at every cutoff, and Appendix~\ref{app:strata} repeats the comparison by severity: the verdict holds in every band of damage.

\section{Which decisions break first, and where the break moves}
\label{sec:breaks}

The surviving fraction varies with bit-width, kind of decision, model and what else is quantized, so it has to be measured on the model being shipped. This section says which decisions go first (\S\ref{sec:dissociation}, Table~\ref{tab:decisions}), where the breaking point moves (\S\ref{sec:secondaxis}) and what repairs the damage (\S\ref{sec:repairs}). Benchmark accuracy alone does not show these thresholds (Appendix~\ref{sec:blind}).

\begin{table}[t]
\caption{\textbf{Every kind of decision at 3 bits, under four quantizers.} Each cell gives two medians: the surviving fraction $c$ over the models where it can be estimated, and the share of decisions that change sign (\%) over all models tested. A superscript gives how many models were measured on that decision when fewer than 16 were. A dash means fewer than four models give a $c$ with standard error at most 0.10. GGUF margins are read against the same model's f16 conversion. Activation-aware scaling is scored on the items not used to calibrate it. The arithmetic row uses 300 items. BFCL and MetaTool rows are public tasks (Appendix~\ref{app:setupfull}).  Appendix Table~\ref{tab:decisionsfull} gives 4, 3 and 2 bits.}
\label{tab:decisions}
\begin{center}\footnotesize
\setlength{\tabcolsep}{3.5pt}
\begin{tabular}{@{}l*{8}{>{\centering\arraybackslash}p{3.45em}}@{}}
\toprule
 & \multicolumn{2}{c}{Round-to-nearest} & \multicolumn{2}{c}{Activation-aware} & \multicolumn{2}{c}{GPTQ} & \multicolumn{2}{c}{GGUF iq3\_xxs} \\
\cmidrule(lr){2-3}\cmidrule(lr){4-5}\cmidrule(lr){6-7}\cmidrule(l){8-9}
Decision & $c$ & flip & $c$ & flip & $c$ & flip & $c$ & flip \\
\midrule
\multicolumn{9}{@{}l}{\emph{Tool use}} \\
\quad Whether to call & 0.21 & 47 & 0.30 & 40 & 0.55 & 21 & 0.71 & 10 \\
\quad Which tool, BFCL & 0.25 & 11 & 0.51 & 4 & 0.59 & 2 & 0.70 & 0 \\
\quad Which tool, MetaTool & 0.24 & 16 & 0.51 & 10 & 0.63 & 6 & 0.76 & 5 \\
\quad Argument fill, ours & 0.61 & 0 & 0.86 & 0 & 0.85 & 0 & 0.90 & 0 \\
\quad Argument fill, BFCL & 0.60 & 4 & 0.67 & 1 & 0.80 & 1 & 0.89 & 1 \\
\quad Tool-result use & 0.35 & 0$^{15}$ & -- & 0$^{11}$ & 0.56 & 0$^{11}$ & 0.67 & 0$^{11}$ \\
\multicolumn{9}{@{}l}{\emph{Refusal}} \\
\quad Forced choice & 0.27 & 31 & 0.35 & 15 & 0.55 & 14 & 0.76 & 6 \\
\quad Open-ended opener & 0.40 & 5 & 0.56 & 3 & 0.70 & 3 & 0.77 & 2 \\
\multicolumn{9}{@{}l}{\emph{Knowledge}} \\
\quad MMLU & 0.19 & 36 & 0.37 & 32 & 0.54 & 27 & 0.73 & 18 \\
\quad BoolQ & 0.51 & 22 & 0.55 & 20 & 0.63 & 12 & 0.87 & 6 \\
\multicolumn{9}{@{}l}{\emph{Procedure}} \\
\quad Arithmetic & 0.33 & 2 & 0.44 & 3 & 0.68 & 1 & 0.59 & 0 \\
\quad Code token & 0.94 & 12 & 0.94 & 0 & 0.85 & 12 & 0.98 & 6 \\
\multicolumn{9}{@{}l}{\emph{Social bias}} \\
\quad BBQ & 0.26 & 40 & 0.41 & 32 & 0.52 & 22 & 0.76 & 15 \\
\bottomrule
\end{tabular}\end{center}
\end{table}

\subsection{Whether to call a tool collapses before which tool to call on most models, and we say on which it does not}
\label{sec:dissociation}

At the bit-width where damage begins, the two halves of a tool-calling decision separate. \emph{Which} tool to call moves less, and on the items we wrote hardly at all. Take the 64 runs in Appendix Figure~\ref{fig:dissociation}, the 16 models under each of the four quantizers at 3 bits. In all 64 not one which-tool decision changes side, against a median whether-to-call flip rate of 32\%, and whether-to-call changes side in every one of them. The which-tool test has eight distinct requests, each scored five times against the same pair of options, so a zero rests on eight measurements and not forty. Appendix~\ref{app:protocol} specifies how requests and options are paired. \emph{Whether} to call collapses, mostly toward not calling. Where quantization damages whether-to-call without destroying it, the model errs toward not calling about two thirds of the time: 65 of the 97 such conditions lean toward not calling. Margins on the should-call side change sign at nearly twice the rate of the other side (Appendix Table~\ref{tab:accounting} gives the filter and each tally).

The condition the rest of the paper follows, 3-bit round-to-nearest on Qwen3-4B, is one of the conditions where whether-to-call collapses, flipping 96\% of should-call against 20\% of should-not-call (Appendix Table~\ref{tab:sideflips}; the other twelve models measured the same way are in Appendix Table~\ref{tab:sideflipsfull}). A model that stops calling flips every item it used to call: most should-call items, and the few should-not-call items it called wrongly at full precision. The offset predicts how far a condition leans: the larger the offset next to the spread, the harder a condition leans. Each model leans in the direction of its own fitted offset, and a minority lean toward calling (Appendix~\ref{app:sideflips}).

Scoring the whole vocabulary says what the offset is made of. At 3 bits the token that opens a tool call sits at the very bottom of the vocabulary-wide change in logits on three models. The call token also sits far below the other reserved tokens, and below ordinary tokens matched to the call token on the size of their output-layer row. These vocabulary-wide and matched-token comparisons are reported in Appendix~\ref{app:correlates}. Both comparisons support a penalty on the call token rather than a vocabulary-wide drift or a token-class effect. Appendix Figure~\ref{fig:dissociation} shows whether-to-call and which-tool parting at 3 bits; at 2 bits which-tool fails too under round-to-nearest and activation-aware scaling.

The tool test set carries four kinds of decision, and the contrast so far uses two. At 3 bits, filling an argument and reading a tool result across turns sit at a median of zero, as which-tool does. Argument filling on public BFCL tasks flips a median of 4\% at 3 bits under round-to-nearest. The tool-result rate is read only over the items whose two answer options begin with different tokens. After rewriting the rest so they do and re-measuring, tool-result use still does not flip at 3 bits (Appendix~\ref{app:sideflips}). Two runs outside these 64 do move which-tool on our items at 3 bits, both on Qwen3-4B: round-to-nearest without a zero point, where which-tool flips everywhere, and one of five stochastic-rounding replicates.  On public tasks which-tool does change, but less: at 3 bits under round-to-nearest it flips a median of 11\% on BFCL and 16\% on MetaTool, against 47\% for whether-to-call, and less than whether-to-call on every model except Gemma-4-E4B, on both sets. Appendix~\ref{app:sideflips} gives the four-kind comparison.

Five checks test whether the whether-to-call/which-tool gap is an item or token effect, and Appendix~\ref{app:sideflips} gives each design in full. Within a condition the which-tool items all start with nearly the same margin, and at 2 bits they lose that margin like every other kind. Pairing items across the two kinds on their full-precision margin gives two dozen matched pairs on Qwen3-4B, and at 3 bits whether-to-call changes side on most of them and which-tool on none. Rescoring the same prompts against other text openers leaves the bit-width of the collapse where it was on every model tried, and changes only how large the collapse is. Ten which-tool items whose two tools are near-synonyms have far less margin to spare than ten whether-to-call items whose request needs the tool, and at 3 bits the large-margin whether-to-call items flip while half of the small-margin which-tool items hold, so margin size is not the cause.

Those results come from items we wrote. We therefore put both decisions on the same 400 tasks from the public tool-calling benchmark BFCL \citep{bfcl2024}, with the test fixed before the runs. There whether-to-call is completing the call and which-tool is naming the right tool, and for each we measured the share of tasks the full-precision model gets right and the quantized one does not (Appendix~\ref{app:crossed}). Five
models answer well enough at full precision to be worth quantizing. Of the 400 tasks, 80 are used to fit $c$, $b$ and $\sigma$ and 320 are held out. Every loss below is read on the held-out tasks the full-precision model got right on both the call and the name, 260 to 300 per model. On three of the five, Gemma-3-4B, Qwen2.5-7B and Qwen3-4B, the completed call goes first by a
clear lead at 3-bit round-to-nearest (Appendix Table~\ref{tab:crossed}). The three do not lose the call the
same way: Gemma-3-4B and Qwen3-4B stop opening it, while Qwen2.5-7B still opens it every time and
fails to finish writing it, so on that model the decision held and the format broke. Averaging the five models equally gives a gap of 20.2 percentage points
between the two losses, and resampling whole tasks puts its lower bound at 17.1. At least three models had to show a gap of 15 percentage points or more, and the average over the models had to have a lower bound above 10. Both hold. Two models do not lose the call first, and they set the scope of the claim. Hermes-3-Llama-3.2-3B loses the name first, and most of its lost names go to another listed tool, so its preference
moved rather than its formatting. Qwen3-1.7B loses both at once: nearly all of its output names no tool, the total collapse the other models show at 2 bits. The margins say why the call goes first. On Gemma-3-4B at 3 bits both decisions lose nearly all of their margin while their offsets separate, a penalty on the call token and none on the name, which is the mechanism of \S\ref{sec:attenuation} on external items.

\subsection{Refusals go one bit lower, and activations are a second axis}
\label{sec:secondaxis}

Under round-to-nearest, most models still keep most of their refusals at 3 bits, where whether-to-call has collapsed. We ask for refusals two ways, a forced choice and an open-ended opener. At 4 bits the current-generation instruct models of 4B and above keep their refusal behaviour under both. At 3 bits, on the forced choice (the more sensitive of the two), four of the 16 models (Qwen3-0.6B, Qwen3-1.7B, Qwen3.5-2B and Gemma-3-4B) lose most of the refusals they made at full precision. The other twelve lose at most 38\%. At 2 bits the median model loses half. Appendix~\ref{app:edges} gives the model-by-model refusal results and the generated-response check below. Decoded behaviour follows the margins at 4 bits and runs ahead of them at lower widths. In free completions from Qwen3-4B the refusal rate is unchanged at 4 bits, falls by a third at 3 bits while the first-token margins still prefer refusal, and at 2 bits the text neither refuses nor complies.  \citet{refusaldir2025} report that the internal direction carrying refusal survives quantization at 8 bits.

The advertised bit-width does not fully describe the quantization configuration. Quantizing the activations as well as the weights is a second axis with its own breaking points. Activations are the harder axis \citep{lin2024qserve}, and we add where those points land per kind of decision. The standard fix is a block-diagonal Hadamard rotation applied before rounding, which spreads the outliers that make activations hard to quantize. We call it \emph{the rotation}. Appendix Figure~\ref{fig:actgrid} plots weight width against activation width, five models from four families. Eight-bit activations are effectively free and combine with four-bit weights. Four-bit activations at the linear-layer inputs, under the plain per-token quantizer, destroy the decision margin \textbf{whatever the weights are doing}. The rotation moves that collapse lower rather than removing it. At 4-bit activations, how much whether-to-call margin the rotation brings back depends on the model: about half on Qwen3.5-4B, a fifth on Granite-3.3-8B and a tenth on Qwen3-4B. Appendix~\ref{app:rotation} compares the plain and rotated schemes on each model. Which activations are quantized matters more than how many bits they get. At 4-bit activations with 8-bit weights, quantizing only the KV cache flips 3 to 18\% of decisions on the four of those models where it was measured, on whether-to-call and the forced-choice refusal. At the same widths, quantizing the inputs to the linear layers flips 44 to 84\% and collapses the surviving fraction.

\subsection{What repairs the damage, and what only looks like it does}
\label{sec:repairs}

Two choices about where bits go move the surviving fraction: the width of the \textbf{KV cache}, the
memory an agent carries across a long context, and the rotation of \S\ref{sec:secondaxis}. At equal nominal width the cache is a more sensitive place to lose bits than the weights, though less so than the linear-layer inputs. On the two models we swept fully, 4-bit weights with a full cache leave most of the margin, while full weights with a 4-bit cache remove a fifth to two fifths of it. The cost falls on the decisions weight quantization hits hardest (Appendix~\ref{app:repairgrid}). No cache width below 8 bits is safe for every model (compare \citealp{alignmentcollapse2026}). How much a cache width costs depends on the quantizer, not on the width alone. Symmetric and asymmetric per-token cache quantizers differ by up to 0.54 in the surviving fraction at the same width. A cache result that does not name its scheme is not reproducible.

The label-free repairs we tested are per-item offsets, three definitions of ``important'' weights, leaving the last blocks at full precision, protecting only the attention layers, rounding each weight by the margin's gradient, adding noise before rounding (dithering), and ordering layers by how much the margin depends on each. Each is set against its control in Appendix~\ref{app:ledger}. None wins on a majority of the models it was tried on, and none comes near the 33-point fall in flip rate that one more bit brings to a severely damaged condition. Of the three importance definitions, margin sensitivity nearly doubles the surviving fraction while its flip rate stays put (Appendix~\ref{app:repairgrid}). One class is excluded by the algebra of \S\ref{sec:attenuation}. Any repair of the form $\ell \to \alpha\ell + \beta$ with a single $\alpha>0$ and a single $\beta$ leaves every sign where it was. This algebra rules out temperature scaling, a global bias offset, and the re-fitted logit scale of \citet{bouguerra2026precise}, whatever that does for calibration.

\section{What you have to measure yourself}
\label{sec:forecast}

Everything so far was measured on models we quantized ourselves. A deployer fits $c$, $b$ and $\sigma$ once on the model being shipped instead of reusing ours.

\paragraph{The relation predicts data it was not fitted on.}

\begin{figure}[t]
\begin{center}
\includegraphics[width=\linewidth]{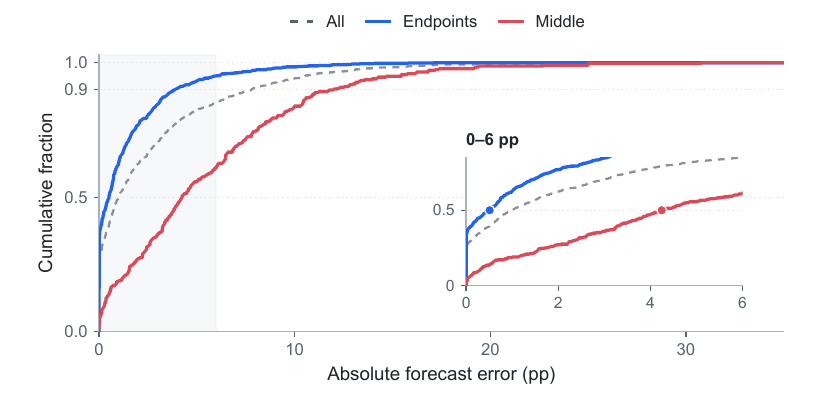}
\end{center}
\caption{Held-out forecast errors across the 1{,}082 conditions. The full CDF includes a magnified view of errors up to 6 percentage points. Circles in the inset mark the medians of the near-endpoint and mid-range conditions. Overall the median and 90th percentile are 1.0 and 7.8 points;
on the 316 conditions whose observed flip rate $r$ lies between 0.2 and 0.8, they are 4.3 and 11.9. The other 766 conditions
lie at or near an endpoint ($r\leq0.2$ or $r\geq0.8$). Their errors are smaller throughout.}
\label{fig:onelaw}
\end{figure}

Fix a kind of decision and a way of quantizing, and suppose the quantized margin is Gaussian around $c\,m + b$ with a spread that does not depend on $m$. The chance that a decision flips is then a normal tail, and with the measured $c$ and $b$ it is
\begin{equation}
P(\mathrm{flip}\mid m) = \Phi\bigl(-(c m+b)\,\mathrm{sign}(m)/\sigma\bigr)
\label{eq:flip}
\end{equation}
with $\Phi$ the standard normal CDF. Fitted on half of a condition's items and scored on the other half, the forecast predicts the flip rate to a median of 1.0 percentage point over the 1{,}082 conditions with enough readable items to split this way, 677 of them from additional experiments. The forecast is least accurate where there is most room to be wrong, on the conditions whose observed rate sits mid-range, as Figure~\ref{fig:onelaw} shows. The forecast also beats simply reusing the fitting half's own flip rate. On the conditions whose observed rate lies between 0.2 and 0.8 the law's median error is 4.3 percentage points against 6.0, and per decision its predicted probabilities sit a fifth closer to what happened, in squared error. Over all 1{,}082 conditions the law is the closer of the two on 47\%, the fitting half's own rate on 33\%, and the rest are level. Appendix~\ref{app:forecastfull} gives the full baseline comparison and checks a split that keeps repeated requests together. Constants copied from another model miss by a median of 26.8 percentage points at 3 bits on the two models tried, against 2.6 at 4 bits. The forecast predicts a change of the margin's sign, not of the model's output. Predicting which calls a model will fail to emit is validated only on Qwen3-4B at 4 bits. On every other quantized model, including every 3-bit one, it does not predict which calls go missing in the BFCL generation test of Appendix~\ref{app:crossed}.

\section{Conclusion}

Which decisions break depends on their kind: at 3 bits under round-to-nearest, whether to call a tool collapses on most models while which tool to call changes less, and most models keep most of their refusals. A deployer can do one paired pass on the model and width being shipped: run the decisions once at full precision and once quantized, and fit the surviving fraction, the
offset and the residual spread to the two readings. The pass forecasts the rate of sign changes wherever the surviving fraction can be estimated. Where it cannot, report the observed sign-change rate.

Four-bit activations into the linear layers destroy the margin whatever the weights keep, and the rotation only moves that collapse lower. A 4-bit cache costs more of the margin than 4-bit weights do. Whether the call collapse holds outside text is open: Appendix~\ref{app:vl} shows why our vision--language test cannot settle it.

\section{Limitations}
\label{sec:limits}
\textbf{What this covers.} Post-training quantization only. Decisions are read at the first token, and how damage accumulates over a rollout is not covered. Sixteen models have their weights quantized. Five have their activations quantized and counted: four of the sixteen and Qwen3-8B, which is on the activation axis only.  One vision--language family covers the second modality. Every weight run quantizes the linear modules inside the decoder blocks and leaves the embedding table at full precision, so a model we call 4-bit is above 4 bits in effective terms. Appendix~\ref{app:setupfull} gives the coverage census.

\textbf{Two limits of scope.} The 4-bit refusal statement holds for standard current-generation instruction-tuned models of 4B and above. Outside that set a model can lose a large share of the refusals it makes at full precision: at 4 bits Qwen2.5-1.5B base loses 43\% of them, and at 3 bits a model can lose all of them (Appendix~\ref{app:edges}). Gemma-4-E4B is left out of the counts on the activation axis because it keeps 35.5\% of its parameters at full precision, which makes its series a weaker intervention than the other four ladder models (Appendix~\ref{app:setuplist}). Its activation conditions still enter the held-out forecast, and dropping them leaves the median error where it is.

\textbf{Mechanism.} We describe the damage (a call-token penalty) but do not explain it, and why large models keep more margin is open.

\textbf{What the external test does not settle.} The break order and the missed-call forecast were tested on one external task set and five models. Four narrower limits are in Appendix~\ref{app:limits}.

\subsubsection*{Ethics statement}
One measurement in this paper is dual-use: 3-bit quantization removes most forced-choice refusals on four of the 16 models and a third of Qwen3-4B's in free generation (\S\ref{sec:secondaxis}). Deliberately harmful
quantization is already published \citep{quantattack2024}, and the removal we measure is what standard post-training quantization does on its own. A precise map of which refusals go first
may still lower the cost of finding them, and we accept that. The margin measurement surfaces the loss of the refusal preference at the first token. The decoded refusal a user reads falls at 3 bits
while that preference still holds on Qwen3-4B, and the responses that lose it mostly deflect the request rather than comply (Appendix~\ref{app:edges}), so first-token preference, decoded refusal and compliance are reported separately. All
models and quantizers are public. The evaluation items, margins and scripts will be released with
the code. The test sets contain no personal data, and no human subjects were involved.

\subsubsection*{AI use statement}
Large language models were used throughout this work, and the disclosure is by category so that
it can be checked. Code: the measurement, analysis and figure scripts were written with model
assistance and reviewed by the authors. Experiments: models helped launch, monitor and collect
the runs. Every measurement is a model forward pass recorded in a named result file. Data: the external tasks are drawn from a public benchmark. The safety requests are XSTest's \citep{rottger2024xstest}, wrapped in prompts the authors wrote; the tool items not from BFCL or MetaTool were written by the authors. Ideas: the form of Eq.~\ref{eq:law} and the hypotheses tested are the authors'; models were used to argue against them and to search the literature. Analysis: models ran much of the analysis and proposed readings of the results, including several that were adopted and several that were wrong. The authors checked each against the stored files and decided what the paper claims. That loop caught two wrong claims and corrected which conditions the headline forecast covers. Writing: models drafted and revised prose, which
the authors edited. Verification: two automated checks run over the source. One lists 422 claims,
each tied to a result file and a phrase in the text, and fails if the stored value is absent
there. A claim not on the list is not checked by it. The other reads the draft against itself: a quantity written two ways in two places, a
figure whose caption and text disagree or that is printed at a scale its labels cannot
survive, a term used before the paper marks its definition. Neither check reads the paper's reasoning, and
model-assisted analysis produced at least one error that reached a draft before a check caught it. The authors have read
every claim that remains and are responsible for all of them.

\subsubsection*{Reproducibility statement}
Every number in this paper is produced by a checked-in script from a named result file. The
listed-claim check is the one described above. Numbers outside its list are checked by hand.
Appendix~\ref{app:setup} gives the item counts, sources and build tags, Appendix~\ref{app:unestimable}
the estimator and its floor, and Appendix~\ref{app:forecastfull} the split protocol. Code, test
sets, and the full result matrix will be released.

\bibliographystyle{iclr2027_conference}
\bibliography{refs}

\begin{thebibliography}{32}
\providecommand{\natexlab}[1]{#1}
\providecommand{\url}[1]{\texttt{#1}}
\expandafter\ifx\csname urlstyle\endcsname\relax
  \providecommand{\doi}[1]{doi: #1}\else
  \providecommand{\doi}{doi: \begingroup \urlstyle{rm}\Url}\fi

\bibitem[Asadi et~al.(2026)Asadi, O'Sullivan, Cao, et~al.]{mirage2026}
Mohammad Asadi, Jack~W. O'Sullivan, Fang Cao, et~al.
\newblock Mirage: The illusion of visual understanding.
\newblock \emph{arXiv preprint arXiv:2603.21687}, 2026.

\bibitem[Ashkboos et~al.(2024)Ashkboos, Mohtashami, Croci, Li, Cameron, Jaggi,
  Alistarh, Hoefler, and Hensman]{ashkboos2024quarot}
Saleh Ashkboos, Amirkeivan Mohtashami, Maximilian~L. Croci, Bo~Li, Pashmina
  Cameron, Martin Jaggi, Dan Alistarh, Torsten Hoefler, and James Hensman.
\newblock {QuaRot}: Outlier-free 4-bit inference in rotated {LLMs}.
\newblock \emph{Advances in Neural Information Processing Systems}, 2024.

\bibitem[Bellam \& Kim(2025)Bellam and Kim]{bellam2025skews}
Abhimanyu Bellam and Jung-Eun Kim.
\newblock Explaining how quantization disparately skews a model.
\newblock \emph{arXiv preprint arXiv:2509.07222}, 2025.

\bibitem[Bouguerra et~al.(2026)Bouguerra, Montoya, Gomez-Villa, Mraidha, and
  Arnez]{bouguerra2026precise}
Aymen Bouguerra, Daniel Montoya, Alexandra Gomez-Villa, Chokri Mraidha, and
  Fabio Arnez.
\newblock Less precise can be more reliable: A systematic evaluation of
  quantization's impact on {VLMs} beyond accuracy.
\newblock In \emph{International Conference on Machine Learning (ICML)}, 2026.

\bibitem[Chang et~al.(2025)Chang, Zhang, Thomason, and Jia]{chang2025breaks}
Ting-Yun Chang, Muru Zhang, Jesse Thomason, and Robin Jia.
\newblock Why do some inputs break low-bit {LLM} quantization?
\newblock In \emph{Proceedings of the 2025 Conference on Empirical Methods in
  Natural Language Processing}, 2025.

\bibitem[Chhabra \& Khalili(2025)Chhabra and Khalili]{refusaldir2025}
Vishnu~Kabir Chhabra and Mohammad~Mahdi Khalili.
\newblock Towards understanding and improving refusal in compressed models via
  mechanistic interpretability.
\newblock \emph{arXiv preprint arXiv:2504.04215}, 2025.

\bibitem[Clark et~al.(2019)Clark, Lee, Chang, Kwiatkowski, Collins, and
  Toutanova]{clark2019boolq}
Christopher Clark, Kenton Lee, Ming-Wei Chang, Tom Kwiatkowski, Michael
  Collins, and Kristina Toutanova.
\newblock {BoolQ}: Exploring the surprising difficulty of natural yes/no
  questions.
\newblock In \emph{Proceedings of the 2019 Conference of the North American
  Chapter of the Association for Computational Linguistics: Human Language
  Technologies (NAACL)}, 2019.
\newblock URL \url{https://arxiv.org/abs/1905.10044}.

\bibitem[Dettmers \& Zettlemoyer(2023)Dettmers and
  Zettlemoyer]{dettmers2023kbit}
Tim Dettmers and Luke Zettlemoyer.
\newblock The case for 4-bit precision: k-bit inference scaling laws.
\newblock In \emph{International Conference on Machine Learning (ICML)}, 2023.

\bibitem[Dong et~al.(2026)Dong, Li, Gao, Li, Zhang, Sun, Chen, and
  Zou]{dong2026hifa4}
Hui Dong, Yanzhao Li, Jie Gao, Chunlu Li, Zhiyuan Zhang, Yupeng Sun, Zhenyuan
  Chen, and Zhiqiang Zou.
\newblock {HiFA4}: Training-free 4-bit {FlashAttention} on {Ascend} {HIF4}
  {NPUs} for {LLM} inference.
\newblock \emph{arXiv preprint arXiv:2607.04302}, 2026.

\bibitem[Dong et~al.(2025)Dong, Tang, Liu, Li, Chu, and Li]{acbench2025}
Peijie Dong, Zhenheng Tang, Xiang Liu, Lujun Li, Xiaowen Chu, and Bo~Li.
\newblock Can compressed {LLMs} truly act? an empirical evaluation of agentic
  capabilities in {LLM} compression.
\newblock In \emph{Proceedings of the 42nd International Conference on Machine
  Learning}, 2025.

\bibitem[Dutta et~al.(2024)Dutta, Krishnan, Kwatra, and
  Ramjee]{dutta2024accuracy}
Abhinav Dutta, Sanjeev Krishnan, Nipun Kwatra, and Ramachandran Ramjee.
\newblock Accuracy is not all you need.
\newblock In \emph{Advances in Neural Information Processing Systems}, 2024.

\bibitem[Egashira et~al.(2024)Egashira, Vero, Staab, He, and
  Vechev]{quantattack2024}
Kazuki Egashira, Mark Vero, Robin Staab, Jingxuan He, and Martin Vechev.
\newblock Exploiting {LLM} quantization.
\newblock \emph{Advances in Neural Information Processing Systems}, 2024.
\newblock arXiv:2405.18137.

\bibitem[Frantar et~al.(2023)Frantar, Ashkboos, Hoefler, and
  Alistarh]{frantar2022gptq}
Elias Frantar, Saleh Ashkboos, Torsten Hoefler, and Dan Alistarh.
\newblock {GPTQ}: Accurate post-training quantization for generative
  pre-trained transformers.
\newblock In \emph{International Conference on Learning Representations}, 2023.

\bibitem[Gao et~al.(2020)Gao, Biderman, Black, Golding, Hoppe, Foster, Phang,
  He, Thite, Nabeshima, Presser, and Leahy]{gao2020pile}
Leo Gao, Stella Biderman, Sid Black, Laurence Golding, Travis Hoppe, Charles
  Foster, Jason Phang, Horace He, Anish Thite, Noa Nabeshima, Shawn Presser,
  and Connor Leahy.
\newblock The {P}ile: An 800{GB} dataset of diverse text for language modeling.
\newblock \emph{arXiv preprint arXiv:2101.00027}, 2020.

\bibitem[Hendrycks et~al.(2021)Hendrycks, Burns, Basart, Zou, Mazeika, Song,
  and Steinhardt]{hendrycks2021mmlu}
Dan Hendrycks, Collin Burns, Steven Basart, Andy Zou, Mantas Mazeika, Dawn
  Song, and Jacob Steinhardt.
\newblock Measuring massive multitask language understanding.
\newblock In \emph{International Conference on Learning Representations
  (ICLR)}, 2021.
\newblock URL \url{https://arxiv.org/abs/2009.03300}.

\bibitem[Huang et~al.(2024)Huang, Shi, Li, Fan, Wu, Zhang, Liu, Zhou, Wan,
  Gong, and Sun]{huang2024metatool}
Yue Huang, Jiawen Shi, Yuan Li, Chenrui Fan, Siyuan Wu, Qihui Zhang, Yixin Liu,
  Pan Zhou, Yao Wan, Neil~Zhenqiang Gong, and Lichao Sun.
\newblock {MetaTool} benchmark for large language models: Deciding whether to
  use tools and which to use.
\newblock In \emph{International Conference on Learning Representations
  (ICLR)}, 2024.

\bibitem[Jang et~al.(2026)Jang, Yang, Lim, and Park]{flatscore2026}
Jiwon Jang, Kisu Yang, Heuiseok Lim, and Hyunwoo Park.
\newblock Flat score, amplified failures: How the error budget masks damage in
  quantized {LLM} agents.
\newblock \emph{arXiv preprint arXiv:2607.27275}, 2026.

\bibitem[Jin et~al.(2025)Jin, Mei, Xu, Sun, Tang, Du, Liu, and
  Zhang]{jin2025massive}
Mingyu Jin, Kai Mei, Wujiang Xu, Mingjie Sun, Ruixiang Tang, Mengnan Du, Zirui
  Liu, and Yongfeng Zhang.
\newblock Massive values in self-attention modules are the key to contextual
  knowledge understanding.
\newblock In \emph{International Conference on Machine Learning (ICML)}, 2025.

\bibitem[Kiselev(2026)]{kiselev2026boundary}
O.~M. Kiselev.
\newblock Boundary-aware quantization: Finite-scale decision geometry of neural
  classifiers.
\newblock \emph{arXiv preprint arXiv:2607.01478}, 2026.

\bibitem[Kumar et~al.(2025)Kumar, Ankner, Spector, Bordelon, Muennighoff, Paul,
  Pehlevan, R{\'e}, and Raghunathan]{kumar2024precision}
Tanishq Kumar, Zachary Ankner, Benjamin~F. Spector, Blake Bordelon, Niklas
  Muennighoff, Mansheej Paul, Cengiz Pehlevan, Christopher R{\'e}, and Aditi
  Raghunathan.
\newblock Scaling laws for precision.
\newblock In \emph{International Conference on Learning Representations}, 2025.
\newblock arXiv:2411.04330.

\bibitem[Li et~al.(2021)Li, Gong, Tan, Yang, Hu, Zhang, Yu, Wang, and
  Gu]{li2021brecq}
Yuhang Li, Ruihao Gong, Xu~Tan, Yang Yang, Peng Hu, Qi~Zhang, Fengwei Yu, Wei
  Wang, and Shi Gu.
\newblock {BRECQ}: Pushing the limit of post-training quantization by block
  reconstruction.
\newblock In \emph{International Conference on Learning Representations
  (ICLR)}, 2021.

\bibitem[Lin et~al.(2024)Lin, Tang, Tang, Yang, Chen, Wang, Xiao, Dang, Gan,
  and Han]{lin2023awq}
Ji~Lin, Jiaming Tang, Haotian Tang, Shang Yang, Wei-Ming Chen, Wei-Chen Wang,
  Guangxuan Xiao, Xingyu Dang, Chuang Gan, and Song Han.
\newblock {AWQ}: Activation-aware weight quantization for {LLM} compression and
  acceleration.
\newblock In \emph{Proceedings of Machine Learning and Systems (MLSys)}, 2024.

\bibitem[Lin et~al.(2025)Lin, Tang, Yang, Zhang, Xiao, Gan, and
  Han]{lin2024qserve}
Yujun Lin, Haotian Tang, Shang Yang, Zhekai Zhang, Guangxuan Xiao, Chuang Gan,
  and Song Han.
\newblock {QServe}: {W4A8KV4} quantization and system co-design for efficient
  {LLM} serving.
\newblock In \emph{Proceedings of Machine Learning and Systems (MLSys)}, 2025.

\bibitem[Nagel et~al.(2020)Nagel, Amjad, van Baalen, Louizos, and
  Blankevoort]{nagel2020adaround}
Markus Nagel, Rana~Ali Amjad, Mart van Baalen, Christos Louizos, and Tijmen
  Blankevoort.
\newblock Up or down? {A}daptive rounding for post-training quantization.
\newblock In \emph{International Conference on Machine Learning (ICML)}, 2020.

\bibitem[Ouyang et~al.(2025)Ouyang, Ge, Hartvigsen, Zhang, Mi, and
  Yu]{ouyang2025lowbit}
Xu~Ouyang, Tao Ge, Thomas Hartvigsen, Zhisong Zhang, Haitao Mi, and Dong Yu.
\newblock Low-bit quantization favors undertrained {LLMs}: Scaling laws for
  quantized {LLMs} with {100T} training tokens.
\newblock In \emph{Proceedings of ACL}, 2025.
\newblock arXiv:2411.17691.

\bibitem[Parrish et~al.(2022)Parrish, Chen, Nangia, Padmakumar, Phang,
  Thompson, Htut, and Bowman]{parrish2022bbq}
Alicia Parrish, Angelica Chen, Nikita Nangia, Vishakh Padmakumar, Jason Phang,
  Jana Thompson, Phu~Mon Htut, and Samuel~R. Bowman.
\newblock {BBQ}: A hand-built bias benchmark for question answering.
\newblock In \emph{Findings of ACL}, 2022.

\bibitem[Patil et~al.(2025)Patil, Mao, Yan, Ji, Suresh, Stoica, and
  Gonzalez]{bfcl2024}
Shishir~G. Patil, Huanzhi Mao, Fanjia Yan, Charlie Cheng-Jie Ji, Vishnu Suresh,
  Ion Stoica, and Joseph~E. Gonzalez.
\newblock The {Berkeley Function Calling Leaderboard} ({BFCL}): From tool use
  to agentic evaluation of large language models.
\newblock In \emph{Proceedings of the 42nd International Conference on Machine
  Learning}, volume 267 of \emph{PMLR}, pp.\  48371--48392, 2025.
\newblock \url{https://gorilla.cs.berkeley.edu/leaderboard.html}.

\bibitem[Proskurina et~al.(2024)Proskurina, Brun, Metzler, and
  Velcin]{proskurina2024confidence}
Irina Proskurina, Luc Brun, Guillaume Metzler, and Julien Velcin.
\newblock When quantization affects confidence of large language models?
\newblock In \emph{Findings of the Association for Computational Linguistics:
  NAACL 2024}, 2024.

\bibitem[Rababah et~al.(2026)Rababah, Qamar, Sparrenberg, Sifa, Kantarcioglu,
  Akcora, and Leung]{rababah2026illusion}
Baha Rababah, Shahzeb Qamar, Lorenz Sparrenberg, Rafet Sifa, Murat
  Kantarcioglu, Cuneyt~Gurcan Akcora, and Carson~K. Leung.
\newblock The illusion of equivalency: Statistical characterization of
  quantization effects in {LLM}s.
\newblock \emph{arXiv preprint arXiv:2607.08734}, 2026.

\bibitem[R{\"o}ttger et~al.(2024)R{\"o}ttger, Kirk, Vidgen, Attanasio, Bianchi,
  and Hovy]{rottger2024xstest}
Paul R{\"o}ttger, Hannah~Rose Kirk, Bertie Vidgen, Giuseppe Attanasio, Federico
  Bianchi, and Dirk Hovy.
\newblock {XSTest}: A test suite for identifying exaggerated safety behaviours
  in large language models.
\newblock In \emph{Proceedings of the 2024 Conference of the North American
  Chapter of the Association for Computational Linguistics: Human Language
  Technologies (NAACL)}, 2024.

\bibitem[Xu et~al.(2026)Xu, Kumarappan, and Zhou]{alignmentcollapse2026}
Bruce~Changlong Xu, Adarsh Kumarappan, and Mu~Zhou.
\newblock Alignment collapse under {KV} cache quantization: Diagnosis and
  mitigation.
\newblock \emph{arXiv preprint arXiv:2606.09864}, 2026.

\bibitem[Zhou et~al.(2026)Zhou, Cao, Li, Yu, Ye, Zhao, and
  Liu]{zhou2026stratified}
Chenxi Zhou, Pengfei Cao, Jiang Li, Bohan Yu, Jinyu Ye, Jun Zhao, and Kang Liu.
\newblock Task-stratified knowledge scaling laws for post-training quantized
  large language models.
\newblock In \emph{Findings of the Association for Computational Linguistics:
  ACL 2026}, 2026.

\end{thebibliography}

\appendix
\section{Appendix}

\subsection{What the instrument reads, and where it stops}
\label{sec:instrument}

The margin instrument scores two designated tokens. Here we compare their preference with
generated responses and identify where the correspondence fails.

\paragraph{The worked example in its population.}
The whether-to-call set on Qwen3-4B has 80 should-call items and 80 should-not-call ones.
At full precision the margins have the intended sign on 77 of the 80 should-call items and on 64 of the 80 should-not-call ones.
The
should-call median is $+18.75$ at full precision, $+13.84$ at 4 bits and $-11.56$ at 3, while the
should-not-call median stays negative throughout. The following transitions use the full-precision-correct items. Counting how many of the 77
correct calls changed side: 4 at 4 bits, then all 77 of them at 3, which is the 96\% of the 80
items quoted in the text, and none of the 64 correct should-not-call margins crosses at either width. At 2 bits
56 of the 77 change side, and 18 of the 64 correct should-not-call margins cross the other way, because there the margins are too
small for their sign to mean much. Round-to-nearest replaces each weight with the nearest value
its smaller number format can represent. Table~\ref{tab:behaviour} gives what the builds emit. Item 54 itself reads $+19.00$ at full precision. At 4 bits it is $+16.00$, which is 84\% of what it was, and at 3 bits $-8.31$.

\paragraph{Overview rendering details.}
\label{app:overview-rendering}
Figure~\ref{fig:overview} uses Qwen3-4B-Instruct-2507. The weight panels show magnitudes of 64 weights from one group in the released checkpoint, ordered by full-precision magnitude; the quantized values use the same order. The blue surface interpolates the original magnitudes as a visual guide, while the red steps show their rounded values. This group illustrates rounding and does not locate the cause of the preference reversal. The probability bars use the Paris request, item 54: the margin between the tool-call opening token and the text-opening token \texttt{I} changes from $+19.00$ to $-8.31$. Probabilities are a softmax at $T=1$ over this pair, with $p(\mathrm{Call})=\operatorname{sigmoid}(m)$, not over the whole vocabulary. Each panel labels its taller bar, and bars near zero are drawn at their true height. The response boxes are illustrative continuations. Actual call emission is measured separately below. Emitting a call does not establish its execution.

Five checks, read from the top of the bit ladder down.

\textbf{At 4 bits it reads behaviour.} The model's own most likely token over the whole vocabulary
is one of the two we score (we call this share \emph{anchoring}) 78\% of the time on Qwen3-4B, Qwen3-1.7B, Qwen2.5-14B and
Granite-3.3-8B, pooled over the tool items. On the open-ended safety opener, a different kind and a
single model, Qwen3-4B refuses in 64-token completions on 70 of the 100 items that ask to be refused
at 4 bits against 73 at full precision, while its margins prefer refusal on all 100 at both widths. Where the two diverge the margin moves the slower, so refusal damage read from margins is a
\emph{lower} bound on what a user would see.

\textbf{At 3 bits it reads a preference; agreement with emitted calls depends on the test set.} That
share falls to 49\% pooled over the same four builds and, on whether-to-call alone, to
30\% across them and 1.9\% on Qwen3-4B, so the margins carrying our headline result are mostly a choice between two designated continuations rather than an emitted one. Decoded freely, with no
constraint to the pair, the Qwen3-4B runs reproduce the direction of the margin result: full precision emits the call
on 77 of its 80 should-call items and answers in text on 68 of its 80 should-not-call ones, while
3-bit round-to-nearest emits the call on none of the 160 while converting none of the 68. Every
build in Table~\ref{tab:behaviour} moves the way its margins do. These decoded counts are not the
margin counts and need not coincide: read from margins the same full-precision model has 64
correct should-not-call margins rather than 68, because on some items the margin prefers the call token while the text
the model actually produces is not a call. The check is not confined to that model. Granite-3.3-8B often writes its call as a fenced JSON block instead of
opening with its call token. At full precision it opens 16 of its 80 should-call items with the call token, and 3-bit round-to-nearest leaves it 2, while on the should-not-call side its replies that do not open with the
call token rise from 62 to 79 of 80. What quantization takes is the call. Its full-precision margins prefer calling on 74 of the 80, more than the 16 it opens with the call token,
so the collapse the margin reports is an upper bound on how many calls opened with the call token were there to lose.

\textbf{The loss is not the model coming apart.} Not calling and not working are different
failures. Of the 3-bit round-to-nearest replies, 22\% switch script and 23\% run digits, against
4\% and 1\% at full precision, so that build is partly broken. The exported 2-bit GPTQ build
stops the same 77 calls with neither pathology and declines in fluent English (\emph{``I'm not able
to provide a function''}), so incoherence is not what produces the collapse.

\textbf{The loss is not an artifact of how the question is put.} The forced-choice safety
instrument reads one letter, and on every item declining is B and listed second, so a build
drifting toward a letter position would look the same. Reversing both, declining first and called
A, moves the flip rate by 1.5 percentage points at 4 bits and 1.0 at 3, .080 to .065 and .275 to
.265 over the same 200 items.

\textbf{At 2 bits it reads nothing.} The fitted line is flat, the surviving margin no longer
depends on the original, and the two option orders part company, .610 against .305. We treat that
as erasure and the end of the measurement rather than a smaller version of the same effect, and
every 2-bit number in this paper is reported under that caveat.

The full-vocabulary check, whether the model's most likely next token is one of the item's two
options (above), lines up with that and adds something the
two-kind version cannot say. Two of the three kinds that hold are also the ones whose margins
still name the token the model would emit (pooled over four builds): 100\% of tool-result items and 93\% of argument-fill.
Which-tool is the exception and it matters: its decisions hold, with no flip at 3 bits on any model in Table~\ref{tab:decisionsfull},
while its anchoring does not, at 33\% against whether-to-call's 30\%. So which-tool survives as
a preference at 3 bits, and the other two survive as behaviour. What survives at 3 bits is therefore not
surviving only as a preference: two of the four kinds keep their anchoring, which the two-kind
contrast on its own cannot show, since both of its halves have lost most of theirs by then.

\subsection{The behavioural check of the margin counts, as a table}

\begin{table}[H]
\caption{The behavioural check of the margin counts: every whether-to-call item decoded freely on
each build, no constraint to the token pair. Full precision emits the call on 77 of the 80
should-call items and answers in text on 68 of the 80 should-not-call ones, and the counts are taken
over those: \emph{stopped} is how many of the 77 a build no longer calls, \emph{started} how many
of the 68 it converts. At 4 bits the three 4-bit builds keep almost every call, 68, 73 and 75 of the 77. The collapse, where a build produces one, is one-directional. These are not
the sign changes of Table~\ref{tab:sideflips}: the 2-bit round-to-nearest build emits neither designated token, so it
stops all 77 calls while only 56 of its 80 margins change sign.}
\label{tab:behaviour}
\begin{center}\footnotesize
\begin{tabular*}{\linewidth}{@{\extracolsep{\fill}}lrrlrr@{}}
\toprule
Build & stopped & started & Build & stopped & started \\
\midrule
GGUF q8\_0 & 0 & 0 & GPTQ w4 & 2 & 5 \\
GGUF iq3\_xxs & 0 & 4 & GGUF iq2\_m & 10 & 5 \\
GGUF q4\_k\_m & 4 & 1 & \textbf{RTN b3} & \textbf{77} & \textbf{0} \\
GPTQ w3 & 2 & 11 & \textbf{RTN b2} & \textbf{77} & \textbf{0} \\
RTN b4 & 9 & 0 & \textbf{GPTQ w2} & \textbf{77} & \textbf{0} \\
\bottomrule
\end{tabular*}\end{center}
\end{table}

\subsection{The forecast: why a Gaussian, and the in-sample check}

A constant-width Gaussian is also the natural guess. A margin is a sum of many weakly dependent contributions, and the spread of such a sum does not follow its mean.

The circular version of Figure~\ref{fig:onelaw}, fitted and scored on all of a condition's items, gives 0.7 points; the held-out version is the one plotted, to avoid overstating accuracy. By what is quantized and how, the held-out error runs from 0.6 points on the key-value cache to 1.1 on activation-aware scaling. Each condition is scored with its own constants, so this checks the shape of Eq.~\ref{eq:flip}, not whether the constants travel; Appendix~\ref{app:forecastfull} tests that.

So the margin fails as a \emph{guarantee}, since under shrinkage no size is safe by itself, and works as a \emph{predictor} once the shrinkage has been measured.

\subsection{Estimator uncertainty and the noise floor}
\label{app:unestimable}

\begin{figure}[H]
\begin{center}
\includegraphics[width=\linewidth]{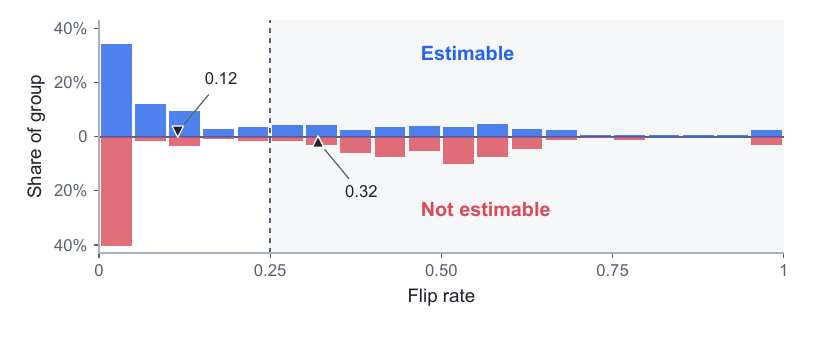}
\end{center}
\caption{Conditions whose slope cannot be estimated still carry substantial damage.
They make up 25\% of the matrix: median flip rate 0.32 against 0.12 for the estimable conditions, with 51\% losing more than a quarter of their decisions against 37\%. They are reported as flip rates rather than
omitted. The mirrored histograms use common 0.05-wide bins and scales, normalized separately within
each group (887 estimable, 296 not estimable). Triangles mark medians; the shaded region marks the strict flip-rate cutoff
above 0.25 used for the stated shares.}
\label{fig:identifiability}
\end{figure}

\paragraph{Conditions whose slope cannot be estimated.}
When the full-precision margins vary too little, we cannot reliably estimate how much they shrink. We require the slope's standard error to be at most 0.10. A quarter of the main matrix's conditions do not meet this criterion. We retain them and report their flip rates: they flip more than the rest, at a median of 0.32 against 0.12. Figure~\ref{fig:identifiability} compares the two groups.

\paragraph{Estimator uncertainty and stochastic-rounding replicates.} The $\pm$ values on $c$ are regression standard errors, given one particular quantization. Applied throughout: \textbf{differences in $c$ below 0.04 are not interpretable.} The regressor is deterministic, so run-to-run variation exists only on the quantized side, where the replicates bound it. Its bf16 values differ in sign from an f16 reference on 3.4\% of Qwen2.5 items and 0.5\% elsewhere (Appendix~\ref{app:gptqchecks}), a small errors-in-variables term that can only pull $c$ toward zero.
\subsection{The rival comparison, in full}
\label{app:rivals}

\paragraph{The four accounts, defined.}
A0 is constant noise ($m'=m+\varepsilon$); A1 the same with a drift term ($m'=m+b+\varepsilon$); A2 margin-scaled noise ($\mathrm{sd}(\varepsilon)=s_0+s_1|m|$); and M is Eq.~\ref{eq:law}. A0 and A1 are special cases of M; A2 is not. Multiplicative against additive is not a distinction of algebraic form, since Eq.~\ref{eq:law} can be rewritten as an additive error $(c-1)m+b+\varepsilon$ whose size grows with the margin. What the comparison rejects in the \emph{core kinds}, whether-to-call and the two ways of asking for a refusal, is a conditional mean fixed at the original margin, with or without a constant offset, even when residual spread is allowed to grow with the margin. At the 2-bit end a full-vocabulary check agrees that the measurement has stopped meaning much: the distribution over next tokens is much more diffuse, at a median entropy of 8.0 nats against 0.01 at full precision.

\begin{table}[t]
\caption{GGUF builds (Qwen3-4B), all four quantized here from one f16 conversion: benchmark score against measured decision flips. Accuracies (\%) are \textbf{BFCL v4} categories, function-calling mode throughout, scored through \texttt{llama-server}: \texttt{simple\_python}, \texttt{irrelevance}, \texttt{live\_simple} and \texttt{multi\_turn\_base}. The version matters: the benchmark's composition changes between releases, so a category score is not comparable across them. Flip shares are against the q8\_0 reference, over the 226 of the 280 tool items whose two scored tokens are among the 20,000 likeliest that llama.cpp returns on all four builds and over the 106 whether-to-call decisions among them. The two flip-share columns share a numerator: every flip counted here is a whether-to-call flip and none of the other 120 decisions changes side on any of these builds, so the columns are one count over two denominators and not two measurements. The multi-turn column is scored on one machine for the top two builds and on another for the bottom two; every accuracy here is read from the scores BFCL's own scorer wrote.}
\label{tab:benchgrid}
\begin{center}
\footnotesize
\begin{tabular}{lcccccc}
\toprule
Build & simple & irrel. & live-simple & multi-turn & flips (all) & flips (call) \\
\midrule
q8\_0 (ref.) & 95.0 & 89.2 & 78.7 & 25.5 & \textendash & \textendash \\
q4\_k\_m & 95.8 & 89.2 & 81.4 & 24.5 & .022 & .047 \\
iq3\_xxs & 88.0 & 85.8 & 72.5 & 15.0 & .027 & .057 \\
iq2\_m & 87.5 & 88.3 & 68.2 & 7.5 & \textbf{.097} & \textbf{.208} \\
\bottomrule
\end{tabular}
\end{center}
\end{table}

\begin{figure}[t]
\begin{center}
\includegraphics[width=\linewidth]{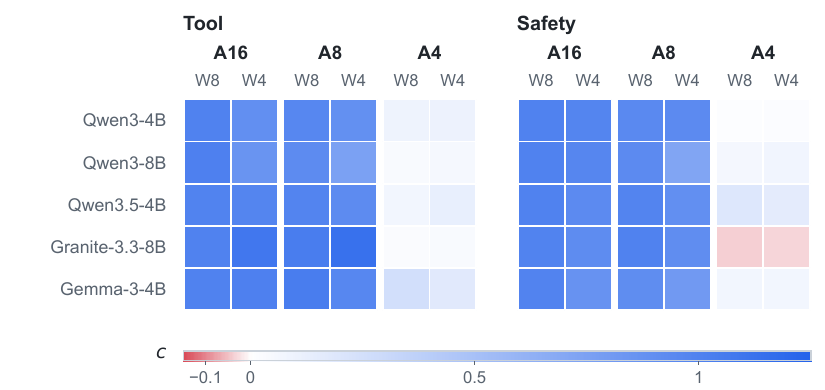}
\end{center}
\caption{The activation-axis grid: 60 plotted conditions, each pooled within a test set
rather than fitted per kind of decision; five models, both test sets (a sixth tag repeats one
model on a second machine and is held out as a replication check). Tool and safety matrices
share one color scale for the surviving fraction $c$, with paired W8/W4 weight columns at each
activation precision A16/A8/A4. White marks zero; negative values are red. At four-bit
activations the weight setting stops mattering; the activation axis dominates the weight axis.}
\label{fig:actgrid}
\end{figure}

\paragraph{Flexible rivals rerun.}
A rerun with more flexible competitors added strengthens the result rather than weakening it. The competitors are an affine form with margin-growing noise, a power-curve mean, their combination, and the union of the two charitable additive rivals, an additive mean carrying both a drift and margin-scaled noise, which grants the additive account every allowance the other two grant separately. That last one wins nothing: over the 44 estimable Qwen3-4B conditions of this rerun it takes no condition at all, and the multiplicative mean beats the best additive rival in 34 of 44 whether or not it is in the pool. Over the 24 damaged conditions \emph{no} winner has an additive mean. One condition deviates from linearity: the safety forced choice at 3 bits prefers a mean that bows upward (a convex power form) by 28.8 BIC over the law, a different condition from the $+28.8$ below, which is the law's \emph{smallest} margin over the additive accounts; the other eight damaged core-kind conditions are won by the law outright. Appendix Table~\ref{tab:adjudication} gives the pooled view across three quantization methods (pooling is the weaker unit of analysis, which is why the per-kind comparison above is the one we rely on); the four-account comparison, run separately on safety-refusal margins across four models and on social-bias margins, reaches the same result wherever damage is real. Model comparison of this kind selects the best predictive description among the stated alternatives; it does not prove a generative mechanism, and the competitor set, though built adversarially, is finite.

\paragraph{The four-account verdict, in full.}
The comparison is run \emph{per kind of decision}, not pooled, because kinds of decision differ in mean margin and a pooled slope can read between-kind differences as shrinkage. Where damage is real (fitted $c<0.70$), the multiplicative account wins every condition, in the kinds the paper's claims rest on (the core kinds) and elsewhere; per-condition BIC differences in the core kinds run $+28.8$ to $+1031.2$ (above 10 is conventionally decisive), and it stays the winner at every cutoff from 0.50 to 0.90 (97.9\% to 100\% of the damaged conditions). That choice of cutoff still selects on the fitted slope, which is the quantity separating the accounts, so the comparison is also run with no such filter at all. Over every fit, grouped by the nominal bit-width the experimenter set rather than by anything estimated, the multiplicative account wins 77.5\% of 2{,}598, and the share tracks the ladder: 49.3\% and 57.3\% at 6 and 5 bits, where little is damaged and the accounts nearly coincide, against 87.1\% at 3 bits and 95.0\% at 2. Nor does the verdict depend on the worst-damaged settings, where the measurement is weakest: stratified by axis and bit-width the 367 core conditions are unanimous in every band, including the 48 in the two bands that keep 79\% or more of the full-precision anchoring (Appendix~\ref{app:strata}).

\paragraph{Where the spread is not constant, measured.} Constant spread is rejected in 38\% of code-token conditions against 15\% for whether-to-call, 9\% for argument filling and 8\% for tool-result use, and on code tokens a constant-spread forecast misses by a median of 6.3 points (4.2 with a spread that grows with the margin) against under 1 point on those three. Weighted least squares shifts $c$ by a median of 0.004.

\subsection{Measurement setup in full}
\label{app:setup}
The four test sets cover 1{,}092 items: tool calling, safety refusals, general knowledge and procedure, and social bias. We built them from BFCL, MMLU, BoolQ and BBQ, and wrote our own items where no public source exists. The first axis is how we quantize the weights, and we do it four ways: round-to-nearest and an activation-aware scaling, both applied in place; calibrated GPTQ builds exported and loaded like any released model; and llama.cpp GGUF builds. Quantizing the activations as well is a second axis. Appendix~\ref{app:setuplist} sets out both.

We fit the relation separately for each run, each bit-width and each kind of decision, and separately again for each side of the three kinds that have one. There are 1{,}183 such fits over 759 run, kind and bit-width groups: 424 of the fits are the two sides of the 212 groups that have sides. Eighteen of the fits come from the first safety test set, which we rewrote before reading any number from it; each condition is fitted on its own items, so the two sets do not have to match, but a comparison across them is not like for like. The significance tests run over 2{,}007 conditions drawn from every run, which is a wider set than the matrix. The forecast is scored on a different set: 1{,}082 conditions that have enough readable decisions to fit on half and score on the other half, drawn from a wider set of runs. Appendix Table~\ref{tab:accounting} gives the source and filter behind each of the main denominators.

One property of the fit matters enough to state here, because it decides what the tables can hold. The surviving fraction $c$ is a regression slope, so it needs a condition's margins to vary; a kind of decision won by about the same amount everywhere has neither a slope nor an intercept we can separate, so we report those conditions by flip rate alone. That happens in 557 of the 2{,}007 audited conditions, more than a quarter of them, and 296 of the full 1{,}183. They are not the quiet ones: their median flip rate is 0.32 against 0.12 for the conditions we can fit. If we dropped them we would hide the worst damage in the paper, so we keep them and report only their flip rates (Appendix Figure~\ref{fig:identifiability}).

\paragraph{What is a distinct measurement, and what is held out.}
\label{app:protocol}
Three of the author-constructed kinds reuse prompts, so an item is not a trial. Counting distinct
full-precision margins in the reference build, whether-to-call has 124
over 160 items, argument filling 28 over 40, tool-result use 15 over 40, and which-tool 8 over 40:
eight requests carried five times, with the same scored token pair each time, so its items are exact
duplicates and every count for that kind is over eight measurements with standard errors that the
repetition understates. The safety set is 200 XSTest requests read by two instruments, 132 and 154
distinct margins over 200 items each; the public sets (BFCL, MMLU, BoolQ, BBQ, and the seeded
arithmetic) do not repeat. Every fit and every flip rate leaves out items whose full-precision margin is exactly zero, which
have no sign to change: the 15 tool-result items on tokenizers that split a number into digits and the 8 code items, whose two options begin with the same
token, and rare exact ties. Runs that re-measure a condition on another machine are excluded
from both the matrix and the forecast. A separate exclusion,
the depth-truncated runs, which read the decision out after the first 9, 12, 18 or 24 of Qwen3-4B's 36 layers, is a truncated network and a different model rather than different
damage; the matrix and the rival comparison exclude it, and the held-out forecast does not, which is
why 67 of its 1{,}082 conditions are depth-truncated builds. Held out means within a condition, fitting on one half of its items and
scoring the flip rate of the other: the held-out forecast splits odd and even items, and because that puts
copies of one prompt on both sides in the three repeating kinds, \S\ref{sec:forecast}'s comparison
against the fitting half's own rate is also run with items sharing a request or a full-precision margin
kept together. Request identity excludes the scored option pair. Under the request-grouped split no request straddles the halves, and on its mid-range conditions (of 802 scoreable) the law's median error is 2.5
percentage points against 4.6 for the fitting half's own rate; Appendix~\ref{app:forecastfull}
reports the comparison. Analyses are
in sample unless stated otherwise.

\paragraph{Test sets, builds and the estimator, in full.}
\label{app:setupfull}
\textbf{What the weight runs touch.}\quad In most of these models the embedding table also turns the final state into output scores, so that table too stays at full precision and the call token's own row is never rounded; the penalty Appendix~\ref{app:correlates} measures on that token comes from the layers before it. The coverage census counts every \texttt{nn.Linear}, tied table included; with the quantizer's own target set it reaches between 82\% and 98\% of the parameters across three models, and almost all of the remainder is that table, with a small additional share in non-linear-module parameters.

The four text test sets are: tool calling, 280 items (160 whether-to-call drawn from BFCL's live-simple and live-irrelevance splits, \citealp{bfcl2024}, plus author-constructed items for which-tool, argument filling and tool-result use, 40 each; these three kinds repeat prompts, so their items are not independent trials; Appendix~\ref{app:protocol} gives the effective count); safety, 200 requests from XSTest \citep{rottger2024xstest}, each wrapped in a prompt we wrote and read by two instruments, so 400 items: 200 by forced binary choice and 200 scored on the reply's natural first token, the \emph{safety opener}; general knowledge and procedure, 220 items (60 MMLU, 60 BoolQ, 60 synthetic arithmetic, 40 code completions); and social bias, 192 items from BBQ \citep{parrish2022bbq}. Appendix~\ref{app:vl} adds two imaged test sets. Whether-to-call reads the token that opens a call against \texttt{I}: the model's reserved call token where it has one and its chat template renders the tools, and otherwise the opening brace of the JSON call, except on Gemma-3-4B, which writes its calls in a fenced block and is read on the fence. Three templates do not render the tools (Gemma-3, Hermes-3 and Phi-4-mini), so the tools go into the system turn as JSON. Hermes-3-Llama-3.2-3B and Phi-4-mini open their calls with other tokens; read against those, the full-precision model prefers calling on most items that need a text reply (on every one for Hermes-3), so we keep the brace, which separates the two sides. On Phi-4-mini the 3-bit flip rate is 0.37 to 0.48 against its own openers and 0.46 against the brace. The quantizers are round-to-nearest at group size 64 applied in place; GPTQ builds calibrated and exported with AutoRound (4, 3 and 2 bits on all 16 models); and four llama.cpp GGUF builds of each model quantized here from one f16 conversion with the importance matrix of Table~\ref{tab:calib}, their margins read against that f16 conversion (the Qwen3-4B benchmark comparison of Table~\ref{tab:benchgrid} takes \texttt{q8\_0} as its reference). The two decision tables read the arithmetic kind on 300 items instead of these 60, built the same way (the answer is the first digit of the sum of two two-digit numbers; Llama-3.1-8B, Hermes-3-Llama-3.2-3B and Phi-4-mini write such a sum as one token, so for them the options are whole sums); no other analysis uses these 300. For the tables, items given without a chat template begin with the model's begin-of-text token wherever its chat template does. The two tables also read which-tool on public tasks: 200 BFCL v3 live\_multiple tasks and 200 MetaTool tasks (tool selection among the listed tools), each scored like our which-tool items, by the first token of the correct tool's name against the first token of one listed tool whose name starts differently. Argument filling is read the same way on 300 BFCL live\_multiple tasks whose correct call has a string argument: the call is opened with the correct tool and parameter name, and the margin is between the first token of the correct value and of that parameter's value in another task. A condition is \emph{estimable} when its slope carries a standard error of at most 0.10 (Appendix Fig.~\ref{fig:identifiability}). Which conditions are unestimable depends on the kind of decision, not on chance: which-tool is 76\% unestimable, whether-to-call 4\%, the safety opener 2\%.

\paragraph{What was quantized, and how.}
\label{app:setuplist}
Quantization settings:
\begin{itemize}\setlength{\itemsep}{1pt}
\item \textbf{Round-to-nearest}, in place at group size 64: 4, 3 and 2 bits on all 16 models, and 8, 6 and 5 bits on five of them; tool, safety, general and BBQ test sets.
\item \textbf{GPTQ}, a calibrated export made with AutoRound: 4, 3 and 2 bits on all 16 models; the same four test sets.
\item \textbf{Activation-aware scaling}, in place: 4, 3 and 2 bits on all 16 models; the same four test sets, each calibrated on 48 of its own items, taken evenly across kinds and left out of scoring.
\item \textbf{GGUF}, quantized here with llama.cpp: q8\_0, q4\_k\_m, iq3\_xxs and iq2\_m on all 16 models; the same four test sets, and BFCL v4 benchmark scores on Qwen3-4B.
\item \textbf{Activations}, with and without rotation: A16 to A3, 5 on the grid, 4 on the ladder, 3 rotated; tool and safety test sets.
\item \textbf{Public tool tasks}: 200 BFCL live\_multiple and 200 MetaTool which-tool tasks and 300 BFCL argument-fill tasks, under all four weight quantizers.
\end{itemize}
The 16 weight-axis models: Qwen3 0.6B/1.7B/32B and Qwen3-4B-Instruct-2507; Qwen3.5 2B/4B/9B; Qwen2.5-Instruct 1.5B/7B/14B; Granite-3.3-8B; Gemma-3-4B; Gemma-4-E4B; Llama-3.1-8B and Hermes-3-Llama-3.2-3B (both Llama-derived); and Phi-4-mini. The 8-, 6- and 5-bit sweeps cover tool and safety decisions on Qwen3-1.7B, Qwen3-4B, Qwen3.5-4B, Qwen2.5-1.5B-Instruct and Qwen2.5-7B-Instruct; Qwen3-4B also has general-task measurements at these widths. Measured alongside but outside the 16 are four base-model controls (Qwen3.5-4B-Base, Qwen3-1.7B-Base, Qwen2.5-7B-Base, Qwen2.5-1.5B-Base) and, on the activation axis only, Qwen3-8B. The activation grid uses Qwen3-4B, Qwen3-8B, Qwen3.5-4B, Gemma-3-4B and Granite-3.3-8B; the ladder swaps Gemma-3-4B for Gemma-4-E4B, which \S\ref{sec:limits} does not count, and the rotation drops Qwen3-8B. 

\paragraph{What the GPTQ column depends on.}
\label{app:gptqchecks}
Two things could make the GPTQ numbers in Table~\ref{tab:decisions} particular to our setup, the text we calibrated on and the particular quantized model that one GPTQ run produced; a third, the 16-bit precision of the full-precision model we compare against, concerns every column. We tested each.

Changing the calibration text changes the knowledge, procedure and refusal numbers, but not whether-to-call breaking before which-tool. We took four models and quantized them a second time with GPTQ, calibrating on 256 documents of pile-10k, a public sample of the Pile \citep{gao2020pile}, instead of the 61 tool-calling dialogues of Table~\ref{tab:calib} (13--18 of them per model pass AutoRound's 1{,}024-token minimum). Whether-to-call flipped at almost the same rate as before (within 3.1 points), and which-tool still flipped nothing. The knowledge, procedure and refusal numbers did change from their values under the dialogue calibration, by up to 0.62 in $c$, some up and some down.

Quantizing the same model with GPTQ a second time gives slightly different numbers but the same answer on whether-to-call breaking before which-tool. GPTQ is not deterministic: two runs on the same model give two slightly different quantized models. We ran it a second time on five models, on another machine, with the same code and calibration text, and scored both results. In all ten model-and-bit-width pairs at 4 and 3 bits the two runs agreed: which-tool flipped nothing in either, and whether-to-call flipped most among the four tool kinds in both. A single flip rate could still differ between the two runs by up to 13 points (Phi-4-mini at 3 bits: 41\% in one run, 54\% in the other). Each GPTQ number in the tables comes from one run. Phi-4-mini's 3-bit arithmetic was scored on a second run of the same recipe, because the build behind its other 3-bit numbers is on a machine we no longer reach.

The full-precision model we compare against is itself slightly noisy, because it runs in 16-bit floats (bf16), as the quantized models do. On Qwen2.5-1.5B, a float32 run and the f16 conversion agree in sign on 1{,}090 of 1{,}092 items, but the bf16 run differs from both on 73. Across the 16 models, bf16 differs from f16 in sign on 3.4\% of Qwen2.5 items (6.7\% on Qwen2.5-1.5B) and 0.5\% elsewhere. For GGUF we therefore compare against the f16 conversion; the other three quantizers are compared against bf16, so this noise is in their numbers.

\paragraph{Token agreement along full-precision continuations.}
\label{app:token-agreement}
The full-precision model greedily generates sixty continuations. At every generated position we record the gap between its two highest-scoring tokens. The quantized model is then scored on the same prefixes, rather than generating its own continuation. A disagreement means its highest-scoring token differs from the full-precision token at that position. At 4 bits, 9.3\% of positions disagree, and the median full-precision gap is 1.0 logits at those positions against 14.4 where the tokens agree. The gap predicts agreement with an AUC of 0.918. At 3 bits, 38.6\% of positions disagree, with an AUC of 0.816. This check uses the top two vocabulary tokens along a fixed trajectory. It is separate from the designated-token measurement and does not measure divergence between two independently generated continuations.

\subsection{Limitations in full}
\label{app:limits}
\textbf{How much of each model the quantizer reaches.} Counting every \texttt{nn.Linear} with the quantizer's own target set, it reaches 81.9\% of the parameters of Qwen3-1.7B, 90.3\% of Qwen3-4B and 97.5\% of Granite-3.3-8B. Almost all of what is left is the embedding table, which every run keeps at full precision, so a build we call 4-bit is above 4 bits in effective terms.

\textbf{A quarter of the full matrix cannot be fitted} and is reported as flip rates instead. Appendix~\ref{app:unestimable} defines the estimation criterion and shows that these conditions have a higher median flip rate, so this limits what we can model, not what we measured.

\textbf{Multiple comparisons are corrected in one place and not the other.} Benjamini--Hochberg covers the per-condition tests of $c=1$ and $b=0$ across 2{,}007 conditions, not the tallies quoted downstream.

\textbf{Where the measurement stops meaning anything.} A full-vocabulary check locates the boundary. At 4 bits the model's most likely next token over the whole vocabulary is one of the item's two options in 78\% of tool items, pooled over four builds, so the margin still reflects what the model would generate. At 3 bits that share falls to 49\%, and at 2 bits to zero on every build. The pooled figure hides two differences. Across kinds, at 3 bits tool-result items are still at 100\% and argument-fill at 93\%, while whether-to-call is at 30\% and tool selection at 33\%. Across models, the pooled 3-bit share runs from 30\% on Qwen3-4B to 66\% on Qwen2.5-14B, so where the instrument stops reading behaviour is the model's own. So at 3 bits both halves of the dissociation in \S\ref{sec:dissociation} are read mostly as a preference between two designated continuations rather than as free generation. That is what grammar-constrained decoding executes. The same check restricts the activation results further, and marks 93 of the 367 damaged core conditions to be read the way the 3- and 2-bit weight results are (Appendix~\ref{app:anchor}).

\subsection{The edges of the account}
\label{app:edges}
\label{sec:vl}
\label{sec:mechanism}

Three things the account does not reach.

\paragraph{We cannot say why it happens.}
\label{sec:chain}

We measured several internal quantities and report them as correlates only, because no experiment here changes a computation and makes the margin move. One explanation we can rule out is that quantization partly undoes instruction tuning: the offset $b$ is already there in models that were never instruction tuned (Appendix~\ref{app:sideflips}).

\paragraph{We could not tell whether it holds outside text.}

We built the same measurement for Qwen3-VL-4B working from a screenshot, on two test sets: a \emph{grounding} set and an \emph{act-or-not} set. Each had to need its screenshot, or a vision result is a language result in disguise, and each had to keep a margin matching what the model would emit. The grounding set passed both checks. The act-or-not set, the one that would have answered our question, failed both, and failed them on the side that carries the result (Appendix~\ref{app:vl}). \textbf{We therefore do not claim} that the one-directional collapse of \S\ref{sec:dissociation} transfers to the second modality. What the grounding set does support is narrower: the form of Eq.~\ref{eq:law} replicates there, on both of the two damaged conditions it can be
fitted on.

\paragraph{A test set can lie to you, and ours did.}

One more test set, the agent-oversight questions, is text only, given to the three Qwen3-VL models (2B, 4B and 8B): 240 items, 120 asking whether an agent acts without a human watching and 120 whether a grant reaches production, balanced so that answering the same word every time scores exactly chance. Eight bits costs almost nothing. At 3 bits four of the six conditions stop reading the item altogether and answer one word to almost every one of the 120, scoring chance while looking decisive. \textbf{We do not claim} that these models under-report unattended agents: asking the same question with its polarity reversed reverses the asymmetry, and the three models then disagree about which way it points. Without the reversed half we would not have caught it.

\paragraph{The refusal ladder, in full.}
At 4 bits the standard current-generation instruct builds of 4B and above keep their refusal behaviour whichever way we ask, including the exported calibrated build (refuse side $0.000$ on both). At 3 bits, counting the same way, the forced choice loses everything on Qwen3-0.6B, 98\% on Gemma-3-4B, 93\% on Qwen3-1.7B and 83\% on Qwen3.5-2B, against 38\% on Hermes-3-Llama-3.2-3B, 37\% on Qwen3.5-9B, 24\% on Qwen2.5-1.5B-Instruct, 20\% on Qwen2.5-7B-Instruct, 15\% on Gemma-4-E4B, 14\% on Qwen3.5-4B, 13\% on Llama-3.1-8B and on Qwen3-4B, 2\% on Granite-3.3-8B, 1\% on Phi-4-mini and none on Qwen2.5-14B-Instruct or Qwen3-32B. Under stochastic rounding rather than the deterministic rule, the same Qwen3-4B build loses 87 to 100\% across five replicates, so at this width the rounding rule matters as much as the model. Back at 4 bits, outside the set that holds, counting a loss as a refusal made at full precision and not made after quantization, over the refusals made at full precision: Qwen2.5-1.5B base loses 43\% of them, Qwen3.5-2B 14\% and Qwen2.5-7B-Instruct 7\%; inside it, a no-zero-point rounding variant of Qwen3-4B loses 5\%. Qwen3-1.7B-Base makes too few refusals at full precision to score, and loses all 13 of them. One base model acquires 17 refusals it did not make at full precision and loses none, which is why gains are counted separately. Within Qwen3 size orders the amount lost at 3 bits on the more sensitive of the two: 0.93 on the 1.7B, 0.13 on the 4B, 0.00 on the 32B. That last leg is measured on the first safety set, the only round-to-nearest run we have for that checkpoint; a calibrated build of it on the current set agrees at zero. Across the whole roster size does not order the amount: Gemma-3-4B loses almost all of its refusals and Phi-4-mini almost none. At 2 bits the median model loses half, from none on Hermes-3-Llama-3.2-3B and Phi-4-mini to all of Llama-3.1-8B's, and on Granite the two instruments disagree (7\% lost on the forced choice against 56\% on the opener), which we report rather than resolve. A free-generation check validates the measurement against behaviour. On the natural-opener items, 64-token greedy completions deliver explicit refusals at an unchanged rate at 4 bits (70\% against 73\% at full precision), at 49\% at 3 bits while the first-token margins still prefer refusal (the responses that stop refusing deflect the request rather than comply), and at 2 bits the completions are incoherent text that neither refuses nor complies. Where margin and behaviour diverge, the margin is the slower to change: refusal damage read from margins is a lower bound on the behavioural loss, not an upper bound.

\subsection{An overall score does not tell you where to spend the bits}
\label{sec:blind}

Three published observations sit at three points on one curve, each at a bit-width where $c$ has not yet fallen far: flat benchmark scores beside accumulating agent failures, refusal directions surviving 8 bits, and single-turn tool use surviving 4 bits. The three-build comparison in Appendix~\ref{app:benchgrid} gives the same ranking by benchmark score change and by margin flip rate.

Whether-to-call damage cannot be read off a benchmark score, because averaging over decisions that flip sharply looks smooth (Appendix Table~\ref{tab:benchgrid}).

The mismatch has a sharper form on a vision--language agent, where the behavioural rate is directly observable rather than inferred: there a language benchmark registers the damage and still points at the wrong layers. This result needs no margin at all, since it is measured on what the model emits. A deployer checks a quantized build on perplexity, and on a whole-model build that check does work: 3-bit group-64 quantization multiplies perplexity by 2.26 while the share of items producing a parseable tool call falls from 0.910 to 0.065. No whole-model build we ran keeps perplexity flat while its calling collapses, but perplexity still cannot say which layers to protect. We quantized nine neighbouring four-block slices of Qwen3-VL-4B's 36-block decoder, one at a time. Four of them hurt tool calling, with Wilson intervals clear of the baseline; the worst three cost between 0.5 and 3.4\% of perplexity and remove a quarter to a half of the calls. Blocks 24--27 sit between two of them and cost nothing at all. The slice with the highest perplexity cost, blocks 0--3, is only the mildest of the four that hurt calling. On that model, perplexity therefore gave no guidance on which layers to protect. Nine slices are nine uncorrected comparisons, and the claim rests on the size of the gaps rather than on their individual $p$-values.

\paragraph{Which layers cost the score, and which cost the calls.} Table~\ref{tab:vldepth} quantizes four contiguous blocks of the 36-block decoder to 3 bits and leaves the rest at full precision, one slice at a time. Perplexity is worst when the first four blocks are quantized, at 1.21 times the full-precision value, and the tool-call rate there is .800 against a full-precision .910. The slice that costs the most calls is blocks 16 to 19, which halves the rate to .450 while leaving perplexity at 1.01, within a point of untouched.

\begin{table}[H]
\caption{Perplexity and tool-call rate under the same 3-bit slice interventions.}
\label{tab:vldepth}
\begin{center}
\scriptsize
\setlength{\tabcolsep}{4pt}
\begin{tabular}{@{}lrrrrrrrrrr@{}}
\toprule
blocks & base & 0--3 & 4--7 & 8--11 & 12--15 & 16--19 & 20--23 & 24--27 & 28--31 & 32--35 \\
\midrule
perplexity ($\times$fp16) & 1.00 & \textbf{1.21} & 1.08 & 1.02 & 1.02 & 1.01 & 1.01 & 1.01 & 1.03 & 1.13 \\
call rate & .910 & .800 & .945 & .870 & .975 & \textbf{.450} & \textbf{.665} & .910 & \textbf{.640} & .875 \\
95\% CI lo & .862 & .739 & .904 & .816 & .943 & .383 & .597 & .862 & .571 & .822 \\
95\% CI hi & .942 & .850 & .969 & .910 & .989 & .519 & .727 & .942 & .703 & .914 \\
\bottomrule
\end{tabular}
\end{center}
\end{table}
\subsection{The repair ledger}
\label{app:ledger}
Each group names the baseline its entries are measured against: one more bit at a matched byte
budget, the same bytes spent another way, or no byte-matched run at all. Numeric entries are median
shares over the conditions that intervention applies to; the others say what was found instead.

\begin{center}\footnotesize
\begin{tabular*}{\linewidth}{@{\extracolsep{\fill}}lr@{}}
\toprule
Intervention & Drop in flip rate \\
\midrule
\multicolumn{2}{@{}l}{\textit{Against one more bit, at the same byte cost}} \\
\quad Per-item offset, oracle labels & 0.282 \\
\quad \textbf{One more bit} (the baseline) & \textbf{0.330} \\
\quad Per-item offset, per kind on calibration items & 0.000 \\
\addlinespace[2pt]
\multicolumn{2}{@{}l}{\textit{Against the same bytes spent another way}} \\
\quad ``Important'' weights, three definitions & no better than the same bytes spread evenly \\
\quad Full-precision trailing blocks & no better \\
\quad Margin-gradient rounding & no better \\
\quad Dithering & no better \\
\quad Attention-only protection & better in 3 of 7 models \\
\addlinespace[2pt]
\multicolumn{2}{@{}l}{\textit{Ruled out without a byte-matched run}} \\
\quad Layer order by margin sensitivity & its own reverse wins 6 of 16 \\
\quad Post-hoc logit rescaling & 0, by algebra \\
\bottomrule
\end{tabular*}\end{center}

\subsection{Related work}
\label{sec:related}
\label{app:relatedfull}
\label{app:related}

Decision-level evaluation is not new. \citet{dutta2024accuracy} show answers flipping both ways while accuracy holds; \citet{dong2026hifa4} measure the same top-two gap and report it shrinking at 4 bits, and \citet{bellam2025skews} report the logit-scale attenuation qualitatively; \citet{kiselev2026boundary} gives the decision geometry at classifier scale. None gives a rate: how fast the gap closes per bit, which kinds close first, and a forecast that holds out of sample. \citet{chang2025breaks} ask which inputs break and answer with residual-stream magnitudes, and we answer with the margin. GPTQ, AWQ, rotation and the drop below 4 bits are well documented \citep{frantar2022gptq,lin2023awq,ashkboos2024quarot,ouyang2025lowbit}; precision-aware scaling laws \citep{kumar2024precision} predict a bit-width's overall \emph{loss}, ours which decisions carry it. A concurrent agent study \citep{flatscore2026} explains quantized tool-calling failures as noise
flipping the decisions whose margin is already thin: we agree on where and correct how (\S\ref{sec:rivals}). \citet{refusaldir2025} find the refusal direction preserved at 8 bits; we measure the refusal margin instead, and it holds at 4 bits and not below, a loss an adversary can trigger \citep{quantattack2024}. \citet{mirage2026} argue that multimodal benchmarks are often answerable without the image. We ran their check on both of ours before quantizing anything, and one of the two failed it (Appendix~\ref{app:vl}). Closest is \citet{alignmentcollapse2026}. They lose refusals under KV-cache quantization at unchanged perplexity, measure flips per prompt as we do per decision, and find collapse onsets spanning four bits with no universal safe width. Attention-based protection is inconclusive on our own measurement (\S\ref{sec:repairs}). We differ on mechanism: they place the cause in a low-dimensional activation subspace. We find that importance-based protection does no better than the same byte budget spread evenly across the layers (\S\ref{sec:repairs}). Our cache runs give their account something to predict: a quantized cache damages decisions selectively by kind, not uniformly.

\paragraph{Where our claim is not the new part.} \citet{proskurina2024confidence} report
that quantization disproportionately affects samples the full-precision model was least
confident about. That is margin selectivity, published in 2024, and no phrasing in this
paper should suggest we found it. What we add is a rate and a functional form for it.
\citet{rababah2026illusion} propose an agreement metric between base and quantized
predictions that is independent of accuracy and run it from 8 down to 2 bits, reporting
non-linear breakpoints; the decision-level move is theirs as much as ours, and the parts
we can claim are the fitted attenuation, the split between kinds of decision, and the
held-out forecast.

\paragraph{Where the literature disagrees with us.} \citet{jin2025massive} find that
quantizers which ignore the massive values in self-attention lose contextual-knowledge
performance, while quantizers aware of them do not, which makes any claim about
``quantization'' rather than about one rounding scheme conditional on the algorithm. Our
matrix carries an activation-aware quantizer alongside round-to-nearest for exactly this reason,
and the fitted form holds in both. \citet{zhou2026stratified} fit task-stratified scaling
laws and conclude that reasoning is the precision-critical skill. Their axis is a task
taxonomy across benchmarks and ours is kinds of decision inside tool use, so the two
orderings are not comparable, and we do not test theirs: our test sets contain no
multi-step reasoning items that would let us.

\paragraph{The three retrodicted observations, in full.}
Before anything new is measured, margin shrinkage explains three published observations that were reported separately. Benchmark scores staying flat while agents accumulate failures \citep{flatscore2026} is the aggregate projection of margins shrinking beneath per-item decision thresholds; Table~\ref{tab:benchgrid} compares benchmark scores with measured flip shares. Refusal directions surviving 8-bit quantization \citep{refusaldir2025} is the law at bit-widths above the collapse: essentially no refusal flips at 4 bits in the current-generation instruct models of 4B and above, and at 2 bits the median model loses half its refusals, which a robustness account stated without a bit-width axis cannot predict. And single-turn tool use surviving 4 bits in agentic benchmarks \citep{acbench2025} is the whether-to-call collapse sitting one bit below the operating point those benchmarks tested. One law, three prior observations, no per-observation story. The bridge is measured: on the three GGUF builds where we have both the benchmark score and the margin measurement, the benchmark's deltas and the measured share of flipped decisions order identically (the live function-calling benchmark of \citealp{bfcl2024}; benchmark $+2.7$, $-6.2$, $-10.5$ points against flipped shares .02, .03, .10).

\subsection{Benchmark scores against the margin instrument}
\label{app:benchgrid}

Table~\ref{tab:benchgrid} expands the three-build comparison above into the full grid: one model's four GGUF builds, all quantized here from one f16 conversion, scored on four categories of the live function-calling benchmark \citep{bfcl2024} beside the margin instrument's flip shares on the same builds. The single-turn categories fall by up to 10.5 points without announcing a collapse, and irrelevance detection ends within a point of where it started (89.2 to 88.3, dipping to 85.8); the multi-turn category falls from 25.5 to 7.5; and the margin instrument sees the flip mass grow $4.4\times$ over the same span, on both denominators. Every score is read in function-calling mode. Categories scored on two machines are mixed here: where a build was scored on both the single-turn categories agree to 0.5 percentage points, but no build has a multi-turn score on both, so the column carrying the largest drop is the one with no cross-machine check. Table~\ref{tab:calib} documents the custom calibration set behind the calibrated quantization paths.
\begin{table}[t]
\caption{The custom calibration set used for the calibrated quantization paths. General text and tool-calling dialogues are mixed at roughly 60/40 by tokens. The two consumers take different cuts of it: the llama.cpp imatrix path takes the 60/40 mixture, truncated at 2048 tokens, while the AutoRound export path takes only the rendered dialogues that reach 1024 tokens. The script that builds it will be released with the code.}
\label{tab:calib}
\begin{center}
\footnotesize
\setlength{\tabcolsep}{3pt}
\begin{tabular}{@{}R{2.9cm}R{7.1cm}R{3.1cm}@{}}
\toprule
Component & Source & Rendering \\
\midrule
General text ($\sim$60\%) & bartowski \texttt{calibration\_datav3} (multilingual, code, technical; the community-standard imatrix text) & raw text \\
Tool dialogues ($\sim$40\%) & Team-ACE ToolACE and NousResearch hermes-function-calling-v1, 300 dialogues sampled, taken in order until they filled the 40\% share, which took 61 & Qwen ChatML template \\
\midrule
\multicolumn{3}{@{}p{13.1cm}@{}}{\footnotesize Consumers: the AutoRound export path (GPTQ builds; the 13 to 18 of the 61 dialogues that reach 1{,}024 tokens) and the llama.cpp imatrix quantizations (all 1{,}143 mixed documents).} \\
\bottomrule
\end{tabular}
\end{center}
\end{table}

\subsection{The rotation intervention}
\label{app:rotation}

\begin{figure}[H]
\begin{center}
\includegraphics[width=\linewidth]{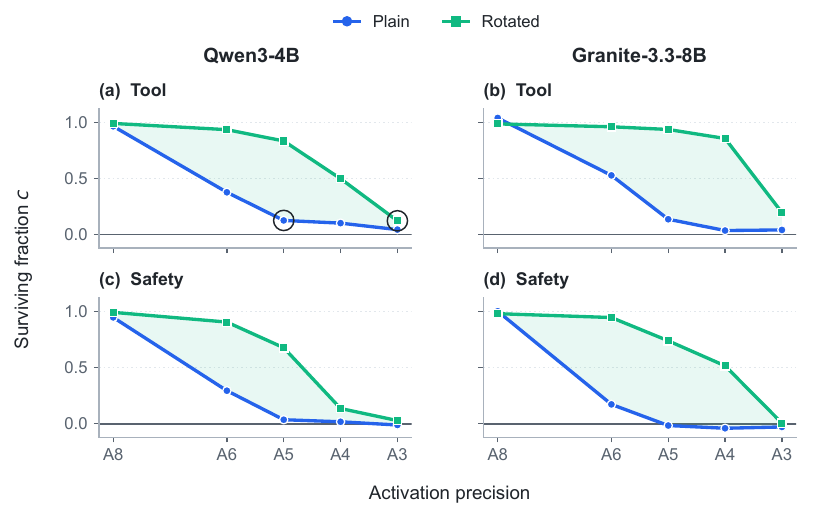}
\end{center}
\caption{The same activation ladder under naive per-token min--max and block-Hadamard
rotation (Plain and Rotated), with weights held at 8 bits. Each point is $c$ over a whole test set. Columns identify models and rows identify test sets; shading separates the measured
trajectories, with true bit-width spacing preserved. Outlined points mark the matched plain
A5 and rotated A3 tool conditions. In both, the should-call side
flips .963 of its decisions against .200 on the should-not-call side, to three decimals apiece;
those are flip rates, not values of $c$, and the agreement is forced by saturation rather than
measured, as the saturation analysis below explains.
The range compression of the worst tensor predicts the shift with a mean absolute error of 0.18 bits on the two models it was developed on and 0.50 bits over all six model-and-test-set conditions.}
\label{fig:rotation}
\end{figure}

\textbf{What the rotation is worth per model.} At 8-bit weights and 4-bit activations into the linear layers, the plain scheme leaves the whether-to-call surviving fraction at or below 0.04 on the four models measured both ways. With the rotation it is 0.11 on Qwen3-4B, 0.21 on Granite-3.3-8B and 0.56 on Qwen3.5-4B, against 1.00 with the activations unquantized; the forced-choice refusal behaves the same way, 0.01, 0.25 and 0.63. The fourth model measured both ways, Gemma-4-E4B, reaches 0.77 on both instruments and is the best result the rotation gives; it is also the model \S\ref{sec:limits} leaves out of the counts on this axis, because it keeps 35.5\% of its parameters at full precision, so it is reported here and not counted in the three above. The rotation is therefore worth about three bits of activation width on average and much less than that on some models, which is why the collapse shift and not a restored margin is what we claim.

\textbf{Weights under 4-bit activations.} Weight precision barely matters once the activations are at 4 bits: on whether-to-call across the five models, dropping the weights from 8 bits to 4 moves the surviving fraction by at most 0.051 and the flip rate by at most 8 percentage points, against flip rates already between 50\% and 84\%.

What the rotation does \emph{not} do is remove the collapse, and that is the part that matters here (Figure~\ref{fig:rotation}). What it does not change is the \emph{kind} of collapse it reaches: both schemes end one-directional toward inaction rather than symmetric, which is not automatic, since round-to-nearest at 2 bits flips both sides (Table~\ref{tab:sideflips}). We claim no more than that from this pair. The two conditions matched here, plain 5-bit and rotated 3-bit activations, both send every whether-to-call margin to the not-call side, so their identical flip rates and flipped items are forced by that saturation rather than measured; and the $c=0.12$ they share is the pooled slope over the whole tool test set, which Appendix~\ref{app:rivals} says not to use.

\paragraph{Rotation bit-widths.}
Those bit-widths are properties of a quantization scheme, and we identify which scheme by measuring an alternative one. The activation quantizer above is per-token min--max with no outlier handling, the setting a substantial line of systems work is designed to escape \citep{ashkboos2024quarot}. Re-running the same runs with a block-diagonal Hadamard rotation around the quantizer moves the collapse \textbf{about three bits lower}: the plain scheme fails between 8-bit and 6-bit activations and the rotated one between 5 and 3, though what whether-to-call keeps at 4 bits differs by model, from 0.11 on Qwen3-4B to 0.77 on Gemma-4-E4B, and in all six model$\times$test-set conditions over three models the 6-bit setting is rescued outright (0.17--0.57 to 0.91--1.03). Interpolating where the surviving fraction crosses 0.70 puts the shift at 2.8 bits. \textbf{The shift is derivable in advance, but less precisely than two models suggested.} A per-token min--max quantizer's step is the token's range over $2^B$, and an orthogonal rotation preserves the norm while shrinking the range, so a range compression of $R$ is worth exactly $\log_2 R$ bits, measurable from forward passes with nothing fitted. On the two models we measured first it works: 2.24 predicted against 2.49 observed and 2.18 against 2.14 (Qwen3-4B, tool and safety), 2.46 against 2.86 and 2.40 against 2.44 (Granite-3.3-8B), a mean absolute error of 0.18 bits. A third model does not follow. On Qwen3.5-4B the predictor gives 1.88 against an observed 2.95 and 2.00 against 3.17, missing low on both test sets and taking the mean over six conditions to \textbf{0.50 bits}. That model's decoder-block linear targets reach 84.9\% of its parameters, with the embedding table and small non-linear modules left at full precision, so coverage alone does not establish comparability. It uses linear attention where the other two use standard attention, a difference the $\log_2 R$ argument does not account for and which we have not tested. We therefore claim the sign and the order of magnitude, enough to plan a bit budget against, and not the precise shift. The predictor must be the 90th-percentile input to the MLP down-projection (\texttt{down\_proj}), not the median tensor (which would predict 0.96 bits): the collapse is set by the worst-conditioned tensor. What the rotation does to that tensor is exactly what ``spreading the outliers'' should mean; its excess kurtosis falls from 141 to 3.

\subsection{The repair grid}
\label{app:repairgrid}

The law says where help must come from, and the two choices that move the surviving fraction are its own suggestions. \emph{What if the bits were spent where the law says?} Moving bits from the weights to the \textbf{KV cache} (the attention keys and values stored and reused during generation) inverts the usual allocation, and a weight$\times$cache grid on two models and both test sets says how far. An 8-bit cache is nearly free: over the 28 checks with both slopes estimable, on two models, five kinds of decision and three weight settings, the median change in $c$ is 0.0055, and only one, tool-result use on Qwen3-4B at 0.068, is above the 0.04 we treat as noise. A 4-bit cache is not, and its damage is selective in the same way the weight axis is, costing a median 0.30 in $c$ on the forced-choice safety test and whether-to-call against 0.11 on the safety opener, argument filling and tool-result use; 3 bits takes the first two to zero. The ladder barely interacts with weight precision, the same shape appearing at 16-, 4- and 3-bit weights as on the activation axis. Four bits spent on the weights leave $c$ at 0.83--1.03 across both models' core kinds; the same four spent on the cache leave 0.57--0.81. Four models at 8-bit weights extend the axis past the grid's two models: Gemma-3-4B, Granite-3.3-8B, Qwen3-4B and Qwen3.5-4B, 18 conditions, and repeat the selectivity at a 3-bit cache: median $c$ 0.09 on the core kinds against 0.73 elsewhere. It also shows what a single collapse point hides. The core-kind slope there spans the full range by model, $-0.01$ on Granite-3.3-8B and $-0.02$ on Qwen3-4B against $+0.83$ on Qwen3.5-4B, so no cache width below 8 bits is safe for every model: \citet{alignmentcollapse2026}'s central observation, reproduced on our instrument and on their axis. One caveat applies to every cache number: the collapse point belongs to the quantizer and not to the bit-width alone. Our two cache quantizers differ only in scheme, per-token symmetric against per-token asymmetric min--max, and they differ by at most 0.041 in $c$ at an 8-bit cache (10 matched conditions) and by as much as 0.536 at 4 and 3 bits (22 conditions). The two-model grids above are the symmetric scheme's and the four-model series is the asymmetric one's, rebuilt from the stored margins; a cache result that does not name its scheme is not reproducible. \emph{What if the outliers were spread before quantizing?} The rotation of \S\ref{sec:secondaxis} is that intervention, and it buys about three bits; the range-based prediction has a mean absolute error of 0.50 bits across six model-and-test-set conditions, with larger errors on Qwen3.5-4B. A third help is narrower: a proof-of-concept activation-aware quantizer (our own implementation, not AWQ, exponent searched on a synthetic probe) helps exactly at the collapse and nowhere else (15 of 19 matched conditions at 3 bits, chance-level at 4 and 2), evidence about this implementation rather than the method class, though the two agree where they can be compared: on Qwen2.5-7B-Instruct at 4 bits a released AWQ build changes the sign of 6 of the 160 whether-to-call margins and none of the 80 on the should-call side, against 10 and 1 for ours; and non-uniform bit allocation across layers helps while the sensitivity \emph{ranking} justifying it does not, ordering by sensitivity performing within noise of its reverse.

\paragraph{The exclusions, in full.}
Against these, the exclusions. One class is ruled out \textbf{by algebra, and it is worth being clear that this is not an experiment}: any post-hoc transform $\ell \to \alpha\ell + \beta$ applying one $\alpha>0$ and one $\beta$ to every logit scales each margin without moving its sign, so temperature scaling and a global bias offset cannot recover a single flipped decision. The scope is that kind and no wider. A correction with a per-class or per-token term, a low-rank output correction, layerwise reconstruction against the full-precision activations, or anything acting on hidden states rather than logits is not covered by the argument and is not tested here; we exclude the transforms whose failure follows from their form, not the idea of post-hoc correction. \textbf{Four fail when measured}, in the specific forms below and at a matched byte budget: protecting ``important'' weights (three definitions, weight norm, input-activation norm and measured margin sensitivity, each protecting nine of Qwen3-4B's 36 layers at a matched 3.25-bit average against the same bytes spread evenly across the layers: on whether-to-call, weight norm and activation norm leave $c$ at 0.16 and 0.16 with flip rates of 0.58 and 0.56, the evenly spread control sits at 0.24 and 0.46, and margin sensitivity raises $c$ to 0.44 while its flip rate stays at 0.43, with the anti-margin control at 0.23 and 0.49), keeping trailing blocks at full precision, margin-gradient rounding, and dithering (noise added before rounding). A fifth, attention-only protection at matched cost, beats protecting the same bytes of MLP weights in only three of the seven models tested (Qwen2.5-7B, Qwen3-4B and Granite-3.3-8B), by more than the 0.04 we treat as noise, and loses by more than that in none, so we do not claim it as a repair. \textbf{One requires knowing each item's correct answer (an oracle), and its label-free variant only diagnoses}: subtracting the per-item measured offset takes a median 0.282 off the flip rate when told which answer is correct, over the 66 conditions where it applies. Estimating $b$ per kind on calibration items and subtracting it, which needs no labels, recovers flips exactly where the law says damage is bias-dominated (35 of 46 on the largest one-directional condition, net $+0.65$) and nets \textbf{0.0 at the median} over the damaged conditions, because where decorrelation dominates there is no bias to remove. The three importance definitions we tried consistently fail against cost-matched controls, which is evidence against those definitions rather than against importance-aware quantization as a research direction. \textbf{One more bit takes a median 0.330 off the flip rate} over the 458 severely damaged conditions, which is the drop in the share of decisions that change sign and not the share of the flips it undoes. Nothing tested \emph{without ground-truth labels} beats it.

\subsection{The measured correlates}
\label{app:correlates}

\paragraph{What the offset is made of.} The margin sees $b$ through two tokens, so a $b$ fitted on one pair cannot separate a drift that moves the whole vocabulary from a penalty on the call token. Recording the change in every vocabulary logit separates them (160 whether-to-call items per model). Both are present and only one of them matters. There is a real vocabulary-wide drift, and its direction is not even shared: at 3 bits the median logit moves $-7.7$ on Qwen3-4B, $-2.4$ on Granite-3.3-8B and $+0.9$ on Qwen3-1.7B. What the three agree on is the call token, which lands at the 0.0th, 0.3rd and 0.5th percentile of that distribution, below almost every other token in the vocabulary, while the alternative it is scored against sits at the 88.7th on Granite. So the offset in Eq.~\ref{eq:law} is not the vocabulary shifting under the pair. It is the call token being singled out, which is why the direction of the collapse is a property of the decision rather than of the model's overall drift.

\paragraph{And not of the token's kind.} The call token is the only one in the test set that is reserved rather than ordinary, so a quantizer that treats rare reserved tokens harshly would put it at the bottom for reasons that have nothing to do with tool calling. Measuring the class alongside it settles which is happening. Two classes are recorded per item: every other reserved token in the vocabulary, and the ordinary tokens whose unembedding row norm is nearest the call token's, which controls for the call token merely having an unusual output row. At 3 bits the class effect exists and is not about being reserved: on Qwen3-4B other reserved tokens fall $-12.9$ against a vocabulary median of $-7.7$, and norm-matched ordinary tokens fall $-12.5$, the same amount. On Qwen3-1.7B neither class falls at all, sitting at the 40th and 38th percentile of a distribution whose median moves up. On no model does the class reach the call token, which is $-2.8$, $-13.1$ and $-8.7$ logits below its own class on Granite-3.3-8B, Qwen3-4B and Qwen3-1.7B and at the 0.3rd, 0.0th and 0.5th percentile against the class's 19th, 8th and 40th. The two effects sized against each other: the norm-matched class falls 1.4, 4.8 and 1.1 logits below the vocabulary median on those three models and the call token a further 2.6, 13.5 and 8.4 below \emph{that} class, so the token-specific penalty is 1.8 to 7.6 times the one norm explains. The triple above is measured against the reserved class and this one against the norm-matched class, which is why the two do not agree to the tenth. It is the larger of the two, and not by a fixed factor.

\paragraph{The repair the offset account predicts, and why it fails.} If quantization pushes one token down by a constant, adding that constant back at inference should undo it, at one float per model and bit-width and no extra bits, and it can be applied after a build has shipped. Swept over the 41 damaged whether-to-call conditions that still have a correct should-not-call item left to lose, the bias that maximises correct decisions is a median 6.5 logits, and it recovers a median 100\% of the lost calls at the cost of a median 100\% of the should-not-call items the quantized build still had right; a bias that costs the decline side nothing recovers anything at all in 12 of the 41, at a median of zero. The two sides score the same pair in opposite senses, and by the width where the penalty matters they sit on the same side of zero, so no scalar separates them. This is the additive counterpart of the algebraic result of \S\ref{sec:repairs}: rescaling the logits moves no sign, and biasing the one token that is singled out moves every sign at once.

\subsection{The second modality, and why it gives no result}
\label{app:vl}

We built the same instrument for Qwen3-VL-4B working from a screenshot, on a \emph{grounding} set
and an \emph{act-or-not} set, and report it here because how it failed is reusable even though what
it found is not.

\paragraph{Two checks a modality test set has to pass.} A vision result that survives without the
image is a language result wearing a screenshot, and a margin that no longer tracks what the model
would emit is not a decision. The grounding set passes both: without the screenshot the model
prefers the target on 179 of 378 items, chance, at a median margin of exactly 0.00, and its
with-image and without-image choices agree on 47.6\%, also chance. The act-or-not set, the one that
would have answered our question, fails both: without the image the model still prefers the target
on 269 of 400 items and agrees with itself 91.8\% of the time, and at 3 bits its argmax over the
whole vocabulary is one of the two we compare on 38.5\% of items, one-sided at 15.0\% of should-act
items, the side carrying the result, against 62.0\% on the other.

\paragraph{What that rules out.} This set cannot say whether the one-directional collapse of \S\ref{sec:dissociation} transfers. The
anchoring failure is one-sided and aligned with the result, so the instrument cannot separate the
asymmetry from its own blind spot. The same check run on text is side-symmetric at 3 bits (Appendix~\ref{app:sideflips}),
which is why the text claim stands and this one does not. What survives is narrower and stated as
such: the form of Eq.~\ref{eq:law} replicates on the grounding set, on both of the two damaged conditions it can be fitted on.

\paragraph{What is not established.} Every model here is Qwen3-VL, since surveying for vision plus
a chat template that accepts tool definitions returns only this one family, so this is a
single-family result. The per-item behavioural forecast of \S\ref{sec:forecast} is not established
here. And the 3-bit results use the same round-to-nearest simulator as the rest of the paper, which is
harsher at 3 bits than a shipping mixed-precision build: our own served \texttt{Q8\_0},
\texttt{Q4\_K\_M} and \texttt{Q3\_K\_M} builds all click within three pixels of one another and all
emit a valid tool call where the simulator parses none. \subsection{How far down the margin still anchors behaviour}
\label{app:anchor}

The full-vocabulary check of Appendix~\ref{app:limits} was also run on the activation axis. Naive per-token activation quantization destroys the anchoring far earlier than weight rounding: 4-bit activations retain 1--3\% of each test set's full-precision share of items whose most likely token over the whole vocabulary is one of the two designated options, against 94--97\% for 4-bit weights and 34\% for 3-bit weights. The rotation brings the anchoring back to 80\% at 4-bit activations and not one bit further down, so it recovers the margin's behavioural meaning over the same range where it recovers any of the margin (\S\ref{sec:secondaxis}). The 93 affected conditions are those at 5-bit activations and below without the rotation, or at 3-bit activations with it, out of 367 damaged core conditions.

\subsection{The one-directional collapse, by condition}
\label{app:sideflips}

\begin{figure}[t]
\begin{center}
\includegraphics[width=\linewidth]{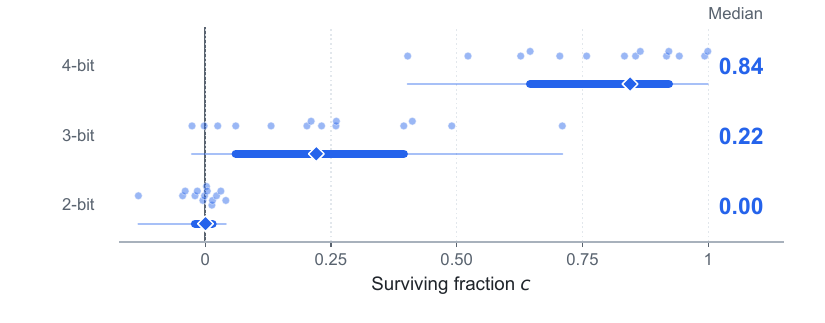}
\end{center}
\caption{The whether-to-call surviving fraction under round-to-nearest, one fit for each of the 14 models whose slope can be estimated at 4, 3 and 2 bits. Each fit contributes one point per
width; vertical offsets separate overlapping points. Diamonds mark medians, thick intervals the
middle half, and thin intervals the full range.}
\label{fig:spread}
\end{figure}

\begin{sidewaystable}
\caption{Table~\ref{tab:decisions} at 4, 3 and 2 bits and the four GGUF builds. Each cell is $c$/flip: the median surviving fraction $c$ over the models where it can be estimated, and the median share of decisions that change sign (\%), over the 16 models. A superscript gives how many models were measured on that decision when fewer than 16 were; -- means that fewer than four models give a $c$ with standard error at most 0.10. GGUF is read against the f16 conversion. Arithmetic uses 300 items. BFCL and MetaTool rows are public tasks.}
\label{tab:decisionsfull}
\begin{center}\footnotesize
\setlength{\tabcolsep}{3pt}
\begin{tabular}{@{}lrrrrrrrrrrrrr@{}}
\toprule
 & \multicolumn{3}{c}{Round-to-nearest} & \multicolumn{3}{c}{Activation-aware} & \multicolumn{3}{c}{GPTQ} & \multicolumn{4}{c}{GGUF} \\
\cmidrule(lr){2-4}\cmidrule(lr){5-7}\cmidrule(lr){8-10}\cmidrule(l){11-14}
Decision & 4 & 3 & 2 & 4 & 3 & 2 & 4 & 3 & 2 & q8\_0 & q4\_k\_m & iq3\_xxs & iq2\_m \\
\midrule
\multicolumn{14}{@{}l}{\emph{Tool use}} \\
\quad Whether to call & 0.86/11 & 0.21/47 & 0.00/54 & 0.86/5 & 0.30/40 & -0.01/54 & 0.88/7 & 0.55/21 & 0.06/49 & 1.00/1 & 0.94/4 & 0.71/10 & 0.56/15 \\
\quad Which tool, ours & --/0 & --/0 & --/65 & --/0 & --/0 & --/50 & 0.62/0 & --/0 & --/0 & 0.99/0 & 0.86/0 & --/0 & --/0 \\
\quad Which tool, BFCL & 0.85/1 & 0.25/11 & 0.03/48 & 0.88/1 & 0.51/4 & 0.00/51 & 0.85/1 & 0.59/2 & 0.17/15 & 1.00/0 & 0.95/0 & 0.70/0 & 0.66/2 \\
\quad Which tool, MetaTool & 0.88/4 & 0.24/16 & -0.01/51 & 0.93/3 & 0.51/10 & 0.00/49 & 0.90/3 & 0.63/6 & 0.19/19 & 1.00/0 & 0.96/2 & 0.76/5 & 0.66/6 \\
\quad Argument fill, ours & 0.92/0 & 0.61/0 & 0.19/51 & 0.95/0 & 0.86/0 & --/48 & 1.00/0 & 0.85/0 & 0.59/0 & 1.00/0 & 1.02/0 & 0.90/0 & 0.83/0 \\
\quad Argument fill, BFCL & 0.92/1 & 0.60/4 & 0.02/50 & 0.94/0 & 0.67/1 & 0.03/50 & 0.93/0 & 0.80/1 & 0.46/10 & 1.00/0 & 0.99/0 & 0.89/1 & 0.83/1 \\
\quad Tool-result use & 0.80/0$^{15}$ & 0.35/0$^{15}$ & --/44$^{15}$ & 0.80/0$^{11}$ & --/0$^{11}$ & --/42$^{11}$ & 0.79/0$^{11}$ & 0.56/0$^{11}$ & --/10$^{11}$ & 1.00/0$^{11}$ & 0.96/0$^{11}$ & 0.67/0$^{11}$ & --/0$^{11}$ \\
\multicolumn{14}{@{}l}{\emph{Refusal}} \\
\quad Forced choice & 0.83/7 & 0.27/31 & 0.00/49 & 0.88/6 & 0.35/15 & 0.00/53 & 0.85/5 & 0.55/14 & 0.01/63 & 1.00/0 & 0.98/4 & 0.76/6 & 0.54/10 \\
\quad Open-ended opener & 0.89/2 & 0.40/5 & 0.01/45 & 0.92/1 & 0.56/3 & -0.01/33 & 0.94/2 & 0.70/3 & 0.09/6 & 1.00/0 & 0.97/0 & 0.77/2 & 0.76/2 \\
\multicolumn{14}{@{}l}{\emph{Knowledge}} \\
\quad MMLU & 0.83/12 & 0.19/36 & -0.01/54 & 0.91/12 & 0.37/32 & -0.02/49 & 0.83/16 & 0.54/27 & 0.00/51 & 0.99/0 & 0.96/8 & 0.73/18 & 0.57/19 \\
\quad BoolQ & 0.89/7 & 0.51/22 & 0.01/50 & 0.93/7 & 0.55/20 & -0.01/53 & 0.91/7 & 0.63/12 & 0.11/38 & 1.00/0 & 0.95/2 & 0.87/6 & 0.75/11 \\
\multicolumn{14}{@{}l}{\emph{Procedure}} \\
\quad Arithmetic & 0.84/0 & 0.33/2 & 0.04/49 & 0.83/0 & 0.44/3 & 0.03/49 & 0.79/0 & 0.68/1 & 0.09/49 & 1.00/0 & 0.96/0 & 0.59/0 & 0.58/0 \\
\quad Code token & 1.00/0 & 0.94/12 & --/50 & 0.97/0 & 0.94/0 & --/50 & 0.96/0 & 0.85/12 & --/25 & 0.99/0 & 0.97/0 & 0.98/6 & 0.91/12 \\
\multicolumn{14}{@{}l}{\emph{Social bias}} \\
\quad BBQ & 0.76/10 & 0.26/40 & -0.01/52 & 0.88/8 & 0.41/32 & -0.01/50 & 0.84/8 & 0.52/22 & 0.04/49 & 1.00/1 & 0.93/5 & 0.76/15 & 0.58/20 \\
\bottomrule
\end{tabular}\end{center}
\end{sidewaystable}

\begin{table}[t]
\caption{Which builds flip the decision to call a tool, and in which direction. Below the rule, five other models at the width
where the damage appears; thirteen models at all three widths are in Appendix
Table~\ref{tab:sideflipsfull}. The two columns are the items whose correct response is a tool call and those whose correct response
is a text reply, 77 to 80 readable items each; GGUF bit-widths are nominal; the ratio divides the tool-call column by the text-reply column. A sign change in either direction counts, so the second column includes items the
full-precision model had wrong and the quantized one gets right. $\dagger$ marks the builds that
send every whether-to-call margin to one side: their ratios are forced by that saturation rather
than measured, and the \S\ref{sec:dissociation} claim about direction counts only the 97 conditions that did not saturate.}
\label{tab:sideflips}
\begin{center}\footnotesize
\setlength{\tabcolsep}{4pt}
\begin{tabular}{@{}llrrrr@{}}
\toprule
Model & Quantizer & Bits & Tool call (\%) & Text reply (\%) & Ratio \\
\midrule
Qwen3-4B & Round-to-nearest & 4 & 5.0 & 7.5 & 0.7 \\
Qwen3-4B & Round-to-nearest & 3 & 96.3\rlap{$^\dagger$} & 20.0\rlap{$^\dagger$} & 4.8 \\
Qwen3-4B & Round-to-nearest & 2 & 70.0 & 40.0 & 1.8 \\
Qwen3-4B & GPTQ (exported) & 4 & 2.5 & 8.8 & 0.3 \\
Qwen3-4B & GPTQ (exported) & 3 & 2.5 & 13.8 & 0.2 \\
Qwen3-4B & GPTQ (exported) & 2 & 96.3\rlap{$^\dagger$} & 20.0\rlap{$^\dagger$} & 4.8 \\
Qwen3-4B & GGUF q4\_k\_m & 4 & 1.3 & 2.5 & 0.5 \\
Qwen3-4B & GGUF iq3\_xxs & 3 & 1.3 & 7.5 & 0.2 \\
Qwen3-4B & GGUF iq2\_m & 2 & 12.5 & 15.0 & 0.8 \\
\midrule
Granite-3.3-8B & Round-to-nearest & 3 & 88.8 & 43.0 & 2.1 \\
Qwen2.5-14B & Round-to-nearest & 3 & 2.5 & 24.1 & 0.1 \\
Qwen3-0.6B & Round-to-nearest & 3 & 32.5 & 66.3 & 0.5 \\
Qwen3.5-2B & Round-to-nearest & 3 & 80.5 & 44.3 & 1.8 \\
Qwen3.5-9B & Round-to-nearest & 3 & 33.8 & 36.7 & 0.9 \\
\bottomrule
\end{tabular}\end{center}
\end{table}

\begin{table}[t]
\caption{Every model that carries both sides of the whether-to-call decision under plain
round-to-nearest, at all three weight widths. Columns as in Table~\ref{tab:sideflips}: the share of
margins that change sign among the items whose correct response is a tool call and among those whose
correct response is a text reply, 76 to 80 readable items of each.
$\dagger$ marks the widths that send every whether-to-call margin to one side, where the ratio is
forced by that saturation rather than measured. The leaning is not shared: at 3 bits seven of the thirteen
flip the tool-call side more than the other and six flip it less.}
\label{tab:sideflipsfull}
\begin{center}\footnotesize
\setlength{\tabcolsep}{4pt}
\begin{tabular}{@{}lcrrr@{}}
\toprule
& & \multicolumn{2}{c}{Margins flipped (\%)} & \\
\cmidrule(lr){3-4}
Model & Bits & Tool call & Text reply & Ratio \\
\midrule
Qwen3-4B & 4 & 5.0 & 7.5 & 0.7 \\
 & 3 & 96.3\rlap{$^\dagger$} & 20.0\rlap{$^\dagger$} & 4.8 \\
 & 2 & 70.0 & 40.0 & 1.8 \\
Qwen2.5-14B & 4 & 1.3 & 5.1 & 0.2 \\
 & 3 & 2.5 & 24.1 & 0.1 \\
 & 2 & 48.8 & 43.0 & 1.1 \\
Qwen3-0.6B & 4 & 35.0 & 27.5 & 1.3 \\
 & 3 & 32.5 & 66.3 & 0.5 \\
 & 2 & 82.5 & 45.0 & 1.8 \\
Qwen3.5-2B & 4 & 6.5 & 32.9 & 0.2 \\
 & 3 & 80.5 & 44.3 & 1.8 \\
 & 2 & 37.7 & 46.8 & 0.8 \\
Qwen3.5-9B & 4 & 5.2 & 8.9 & 0.6 \\
 & 3 & 33.8 & 36.7 & 0.9 \\
 & 2 & 59.7 & 60.8 & 1.0 \\
Granite-3.3-8B & 4 & 5.0 & 16.5 & 0.3 \\
 & 3 & 88.8 & 43.0 & 2.1 \\
 & 2 & 92.5\rlap{$^\dagger$} & 46.8\rlap{$^\dagger$} & 2.0 \\
Qwen3-1.7B & 4 & 6.3 & 24.1 & 0.3 \\
 & 3 & 62.0 & 40.5 & 1.5 \\
 & 2 & 55.7 & 35.4 & 1.6 \\
Qwen3.5-4B & 4 & 10.5 & 11.4 & 0.9 \\
 & 3 & 63.2 & 26.6 & 2.4 \\
 & 2 & 50.0 & 51.9 & 1.0 \\
Qwen2.5-1.5B & 4 & 6.3 & 20.0 & 0.3 \\
 & 3 & 81.3 & 47.5 & 1.7 \\
 & 2 & 60.0 & 46.3 & 1.3 \\
Qwen2.5-7B & 4 & 0.0 & 8.8 & 0.0 \\
 & 3 & 3.8 & 36.3 & 0.1 \\
 & 2 & 92.5 & 38.8 & 2.4 \\
Gemma-3-4B & 4 & 2.5 & 30.0 & 0.1 \\
 & 3 & 68.8 & 75.0 & 0.9 \\
 & 2 & 30.0 & 41.3 & 0.7 \\
Hermes-3-Llama-3.2-3B & 4 & 26.3 & 11.4 & 2.3 \\
 & 3 & 58.8 & 17.7 & 3.3 \\
 & 2 & 72.5 & 13.9 & 5.2 \\
Phi-4-mini & 4 & 38.5 & 25.0 & 1.5 \\
 & 3 & 42.3 & 48.8 & 0.9 \\
 & 2 & 60.3 & 41.3 & 1.5 \\
\bottomrule
\end{tabular}\end{center}
\end{table}

\paragraph{Where each model falls on the way down.} The collapse is universal and its timing is
not. At 4 bits the 14 fits of Appendix Figure~\ref{fig:spread} run from 0.40 to 1.00, with a median of 0.84; at 3, the width where the damage begins,
they scatter widest, from 0.03 below zero to 0.71; and at 2 bits every one is under 0.05 with a
median of 0.00.

\paragraph{What the instrument can and cannot see at 3 bits.} The 3-bit whether-to-call collapse does fix a direction: of the 14 conditions where every decision has already gone the same way, all go toward not calling. At this bit-width the flips are read as conditional preferences, not as emitted tokens (\S\ref{sec:instrument}). So we ran here the check of Appendix~\ref{app:vl}: the share of items whose full-vocabulary argmax stays inside the option pair at 3 bits is 30.6\% on the should-call side and 28.8\% on the other, over four builds, against 77.2\% and 46.3\% at 4 bits. What is nearly even is where the two sides end up: 30.6 against 28.8. What is not even is how far each fell to get there, 46.6 percentage points against 17.5. The asymmetry needs only the even end points, not even falls. At the bit-width where the 4.8-fold result is read, the instrument is equally blind on both sides, so its blindness cannot make one side look worse than the other.

\paragraph{Why whether-to-call breaks before which-tool: the checks in full.} The two kinds come from different items, 160 whether-to-call against eight distinct selections carried as 40 items, so the difference could in principle come from the items. Within any one condition the eight selections are decided by nearly the same margin (median $|m|$ 14.8 logits, within-run spread 0.24 logits, across 144 runs), so no difficulty gradient among them could produce it. The same 40 do collapse one bit lower, so the items are not immune: the kind of decision sets the bit-width of the collapse. A third check swaps which kind has margin to spare, on the same model and quantizer: ten which-tool items whose two tools are near-synonyms, median margin 6.2 logits, against ten whether-to-call items whose request is impossible without the tool, median 22.4. If the difference were distance from crossing, the small margins would flip and the large ones would hold. At 3 bits the large ones flip 10 of 10 and the small ones 5 of 10: giving whether-to-call more room does not save it and taking which-tool's away does not finish it. What differs is the drop itself: $-35.2$ logits for whether-to-call against $-7.1$ for which-tool on the same build, which is the kind of decision and not the item. A fourth pairs the two kinds by margin rather than trusting them to be alike. Within a condition the eight selections carry nearly the same margin, and whether-to-call is spread wide,
so which-tool could survive because its items were written with margins larger than any offset
the quantizer produces. Measuring both kinds at full precision first and pairing items
across kinds to within 1.5 logits gives 24 pairs on Qwen3-4B at medians of 15.03 and 14.94.
At 3 bits, on those pairs, whether-to-call changes side on 17 of 24 and which-tool on none. The pairing is what limits this: on Qwen3-1.7B only 6 pairs
exist and on Granite-3.3-8B none do, because there the two kinds' margins do not overlap at
all, at medians of 3.75 and 13.38. That the two kinds differ in margin is exactly the objection,
and it is answerable only where they overlap.

A fifth check varies the token the decision is read against. Both sides of the kind score the call token against \texttt{I}, so the headline is measured through one alternative on all 160 items, and what collapses could in principle be that token. Rescoring the same prompts against \texttt{The} and against \texttt{Sure} leaves the collapse where it was: on Qwen3-4B the should-call side changes side on 77 of its 80 items at 3 bits against \texttt{I}, on 77 against \texttt{The} and on all 80 against \texttt{Sure}, and on 4, 2 and 2 of them one bit higher. The bit-width of the collapse does not move with the alternative on any model tried. The size of the collapse does: on Granite-3.3-8B the same count is 71 of 80 against \texttt{I} and 53 of 80 against \texttt{The}, so the instrument reads the decision while how much of it moves depends on what the decision is read against.

\paragraph{What the tool-result kind could and could not read.} On tokenizers that split a number into digits, the tool-result rate is read over 25 of that kind's 40 items; Llama-3.1-8B, Hermes-3-Llama-3.2-3B and Phi-4-mini write each two-digit temperature as one token and read all 40. In the other 15 the two temperatures on offer share a leading digit, so the two scored tokens are one token, the margin is identically zero and the item cannot change sign whatever quantization does. No item in the other three tool kinds is like that, and counting the 15 in or out leaves every 3-bit flip median of Table~\ref{tab:decisionsfull} and the zero-flip counts below unchanged. The code-token kind has 8 such items, whose two options both begin with a space, and they are left out the same way. This is repaired in a separate test set that moves the fifteen distractors off the leading digit. Re-measured there, all 40 items are live on all four builds tried, and the flip rate is zero at 4 bits and at 3 on every one, against a median 0.51 at 2. What survives at 3 bits is the kind, not the items that happened to be readable.

\begin{table}[H]
\caption{Where each denominator comes from. ``Estimable'' means the slope's standard error is at most 0.10; ``damaged'' means a fitted $c<0.70$. Each row lists its unit, source and filter. Each analysis uses a different set of conditions, so a count from one row does not apply to another.}
\label{tab:accounting}
\centering
\footnotesize
\begin{tabular}{@{}R{1.7cm}R{3.6cm}R{7.7cm}@{}}
\toprule
count & unit & source and filter \\
\midrule
1{,}183 / 887 & matrix conditions / estimable & main matrix: 86 runs, mostly round-to-nearest and activation-aware, $\times$ kind $\times$ bit-width, intercept estimator; the 296 unestimable conditions are the 25\% described in Appendix~\ref{app:unestimable} \\
2{,}007 / 557 & audit conditions / unestimable & the per-condition tests of $c=1$ and $b=0$ over every run; 27.8\% report only a flip rate \\
1{,}450 & forecast-computable conditions & full-condition fits with se$(c) \le 0.10$ and residual $\sigma > 0$; the held-out forecast is a separate population, half-fit, 12 items a half \\
1{,}438 & constant-spread audit conditions & wider set of runs, $n \ge 24$, estimable \\
1{,}580 & leave-one-out transfer conditions & each predicted from the others' median constants at its bit-width \\
24 & damaged comparison conditions & per-kind model comparison, main example model's conditions, fitted $c < 0.70$ \\
642 / 1916 & damaged / all per-kind conditions & the same comparison over every run \\
111 / 97 / 65 & direction conditions / unsaturated / leaning toward not calling & the direction tally over the whether-to-call runs of the main experiments, seed, activation and depth variants included; the exported builds added for Tables~\ref{tab:decisions} and~\ref{tab:decisionsfull} are not in it. A condition enters with at least 20 paired items per side and a flip rate above 0.20 on at least one side; 14 send every margin to one side and are dropped \\
17 & models in the direction tally & the same 111 conditions counted by model: 13 of the 16 models (Llama-3.1-8B, Hermes-3-Llama-3.2-3B and Phi-4-mini have no run in it), three base models and the activation-only Qwen3-8B \\
49 & offset-law conditions & whether-to-call conditions of the main matrix that carry both sides, flip on each side and do not send every margin to one side \\
48 / 64 & four-kind pairs / pairs with both whether-to-call and which-tool & Appendix Table~\ref{tab:decisionsfull} at 3 bits (iq3\_xxs for GGUF), one per model and quantizer \\
1{,}082 & held-out conditions & Figure~\ref{fig:onelaw}: odd/even halves within each condition, all damage axes; a wider set of runs than the matrix, 677 of them outside it \\
710 & repair-test conditions & calibrated-offset repair: damaged conditions with at least 5 flips \\
458 & repair-baseline conditions & the one-more-bit baseline: matrix conditions with flip $\ge 0.25$ that were also measured one bit higher \\
1{,}409 & clustered-closure conditions & the by-model clustering of Appendix~\ref{app:accounting}: per-condition closure recomputed over every run, four unattributable runs excluded \\
1{,}430 / 193{,}779 & calibration conditions / predictions & per-item flip probabilities: conditions with $n \ge 24$ and se$(c) \le 0.10$ \\
\bottomrule
\end{tabular}

\end{table}

\begin{table}[H]
\caption{Does quantization add noise or shrink the signal? Pooled fits per condition; the per-kind comparison is in \S\ref{sec:rivals}. The first four rows are Qwen3-4B and the last two name their own model. $\Delta$BIC is the law's margin over the best of the three additive accounts, and $>10$ is decisive; a negative entry means one of them fits better. Two entries are negative, GGUF q4 and GPTQ w4, both undamaged ($c$ of 1.01 and 0.98) and within noise. The correlation column is evidence against constant-noise accounts only; $R^2$ is fit quality only. RTN is round-to-nearest.}
\label{tab:adjudication}
\begin{center}
\footnotesize
\resizebox{\textwidth}{!}{\begin{tabular}{lrrrr}
\toprule
Condition & survival $c$ & fit $R^2$ & $\Delta$BIC (shrinkage) & corr($|m|$, $|m'-m|$) \\
\midrule
RTN b4 / b3 / b2 (tool) & 0.90 / 0.47 / 0.02 & .95 / .52 / .01 & $+51$ / $+221$ / $+800$ & $-$.15 / .08 / .93 \\
RTN b4 / b3 / b2 (general) & 0.90 / 0.30 / $-$0.03 & .88 / .13 / .01 & $+11$ / $+92$ / $+493$ & .07 / .55 / .89 \\
GGUF q4 / iq3 / iq2 (tool) & 1.01 / 0.80 / 0.60 & .99 / .94 / .77 & $-3$ / $+63$ / $+85$ & $-$.35 / .23 / .56 \\
GPTQ w4 / w3 / w2 (tool+general) & 0.98 / 0.79 / 0.38 & .95 / .87 / .20 & $-4$ / $+159$ / $+197$ & $-$.10 / .27 / .36 \\
RTN b4 / b3 / b2 (tool, Qwen3-0.6B) & 0.83 / 0.04 / $-$0.03 & .71 / .03 / .01 & $+21$ / $+753$ / $+697$ & $-$.34 / .73 / .81 \\
RTN b4 / b3 / b2 (tool, Llama-3.1-8B) & 0.78 / 0.85 / 0.09 & .93 / .75 / .09 & $+200$ / $+21$ / $+629$ & .64 / .27 / .94 \\
\bottomrule
\end{tabular}}
\end{center}
\end{table}

\begin{figure}[H]
\begin{center}
\includegraphics[width=\linewidth]{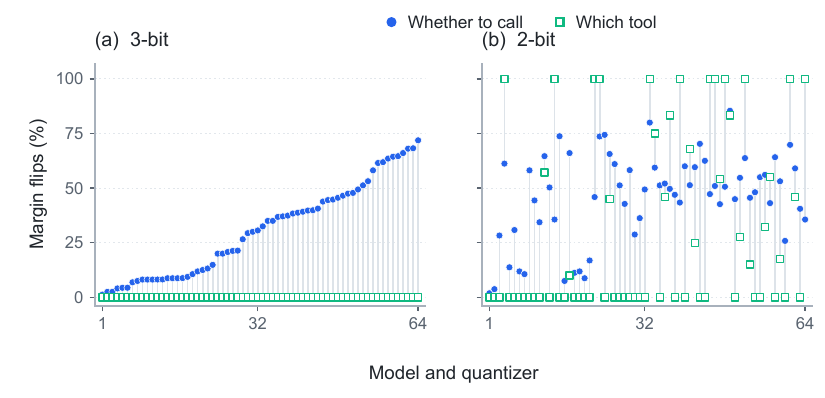}
\end{center}
\caption{Whether-to-call against which-tool in the same 64 pairs, 16 models under each of four quantizers (GGUF at iq3\_xxs and iq2\_m).
Within each column a connector pairs whether-to-call with which-tool, preserving exact zero
rates. The 3-bit and 2-bit panels use the same order, sorted by the 3-bit whether-to-call rate.
Which-tool is untouched in 64 of the 64 pairs at 3 bits. At 2 bits it is untouched in 36 of 64: 30 of the 32 GPTQ and GGUF pairs and 6 of the 32 round-to-nearest and activation-aware ones.
Zero flips do not establish a zero population rate, and the forty which-tool items are eight requests carried five times (Appendix~\ref{app:protocol}).}
\label{fig:dissociation}
\end{figure}

\paragraph{Controls on the direction of the collapse.}
\textbf{The direction tally.} The tally itself selects 111 whether-to-call conditions with a flip rate above 20\% on at least one side, a filter on observed flips rather than on the fitted slope. Of these, 14 have every margin on one side after quantization; such a condition flips whatever the full-precision model got right rather than what quantization did, so they are dropped, and all fourteen saturate toward not calling. Of the remaining 97, 65 lean toward not calling, at median flip rates of 0.68 on the should-call side against 0.38 on the other.

\textbf{The direction, counted by model.} Counted over the 17 models represented in this tally, including base-model controls and the activation-only Qwen3-8B, with each model's median gap taken first, 10 of 17 lean toward not calling, 6 lean the other way and 1 is exactly level ($p = 0.63$ two-sided, $0.45$ without the level model). The directional tendency is therefore not statistically established at the model unit.

\textbf{The offset and the lean.} Over the 49 whether-to-call conditions of the main matrix that carry both sides, flip at least one item on each and do not send every margin to one side, the more-flipped side flips a median 1.8 times as often as the other where $|b|/\sigma < 1$ (37 conditions) and 2.4 times between 1 and 3 (11); the one condition above 3 flips 10 times as often, and the rank correlation across the whole range is 0.52.

\textbf{The four kinds, counted.} Counted once per model and quantizer at 3 bits (Appendix Table~\ref{tab:decisionsfull}), whether-to-call changes side on every one of the 48 model-and-quantizer pairs carrying all four kinds, at a median rate of 0.32, while argument filling and tool-result use show no flip at all on 44 and 42 of them and which-tool on all 48. Across all 64 pairs, whether-to-call flips more than which-tool in every one.

Three things make the one-directional collapse more than a curiosity of one condition. It is \textbf{not an artifact of margin size}. Within bands of the $|m|$ range shared by the two sides (should-call and should-not-call), permuting the side labels gives an exact null with margin size removed by construction; a directional flip-rate gap survives that test in 8 of 10 models at 3 bits (5 toward not calling, 3 toward calling) and 3 of 10 at 2 bits. In all 11 directional cases the gap's direction matches the sign of the model's fitted offset, which was fitted on margins and never on flips. Nor is the pooled line an artifact of the two margin clusters, though the two sides sit in separate ranges of $m$ and a single line could in principle be fixed by the two cluster centres alone: where the pooled slope reports deep damage ($c < 0.30$; 116 side-labelled conditions), the larger of the two within-side slopes has a median of 0.10 and exceeds 0.5 in 1 of the 116 conditions. In the flagship condition the two within-side slopes refit to 0.24 and 0.10, with the same large negative offset. A small pooled $c$ is therefore not hiding an intact side. It \textbf{survives on an exported build}: the exported GPTQ w2 model collapses one-directionally too, while above the collapse (w3, $c=0.81$) it does not lean toward not calling (2.5\% against 13.8\%), the law's own prediction. And what it tracks is \textbf{the offset relative to the residual noise}, not the method. Rank the nine Qwen3-4B builds of Table~\ref{tab:sideflips} by $|b|/\sigma$, which the table itself does not carry: the two at 3.8 and above are the two that flip one-directionally, and every other build is at 0.8 or below, including RTN b2, whose $c$ has also reached zero but whose offset has not. (One whether-to-call slope among the nine sits above 1: GPTQ w4, at $1.03 \pm 0.02$. Mild amplification at shallow damage is a matrix-wide pattern, not a fluke of this build: 15\% of estimable conditions at 4 bits and above sit above 1, with median excess 0.037, under plain rounding and calibrated methods alike. The excess here is within the 0.04 floor of Appendix~\ref{app:unestimable}.) The asymmetry belongs to the early stage of damage, where the offset $b$ outweighs the residual noise; at 2 bits the collapse is no longer one-directional ($c \approx 0$, flips $.700$ against $.400$). The pattern is not simply explained by model size, and we report the counts rather than force a story. Qwen3.5-9B shows no asymmetry at either bit-width. Among the small models, Qwen3-0.6B flips the text-reply side more at 3 bits and the tool-call side more at 2, Qwen3.5-2B does the opposite, and Qwen3-1.7B and Qwen2.5-1.5B flip the tool-call side more at both (Table~\ref{tab:sideflipsfull}).

\paragraph{Base-build control on the offset.}
A tempting account of the offset is reversal: quantization partially undoes instruction tuning, so the model drifts back toward its pretrained prior, which would explain why the drift so often points away from acting. Matched base builds test it, because a model that never received instruction tuning has nothing to undo: the account predicts a near-zero offset, as a fraction of the model's own typical margin, in every base. The four matched pairs at 3 bits on whether-to-call show otherwise. One base's offset is four times the size of its sibling's and of opposite sign (base $-0.84$ against instruct $+0.20$; Qwen2.5-7B); one is comparable to its sibling's ($-0.55$ against $-0.59$; Qwen2.5-1.5B); one is near zero, as the account predicts ($+0.04$ against $-0.34$; Qwen3.5-4B); on one the offset cannot be fitted (Qwen3-1.7B). An offset already present in two of the three bases where it can be fitted cannot be an artifact of undoing tuning; whether reversal contributes in some models stays open.

\subsection{Both halves of a call on the same external tasks}
\label{app:crossed}

\paragraph{Design.} Four hundred BFCL v4 live-multiple requests with one ground-truth call and at
least two tools, the correct tool's position shuffled per task, 80 tasks for calibration and 320
sealed, split by task id. Each task is decoded three ways on
every build, greedy, through the model's own chat template: with its tools, where the model should
call the correct one; with the correct tool removed, where it should not call; and after the call
opener alone, where the complete name is matched against the tool list. The same tasks carry the
whether-to-call pair on both variants and the which-tool pair after the shared prefix of the names,
so margins and emitted decisions are read on one item set. Three templates render the same prompt
with and without the tools argument: Gemma-3, Hermes-3 and Phi-4-mini receive the definitions as
JSON in the system turn instead, and each model's pair uses the token it emits, a fenced block for
Gemma and a brace for the other two. A build enters only if at full precision it opens and names a
call on at least 0.8 of the tasks when nothing forces it to, and names the correct tool on at least
0.8 when the call format is given to it. Granite-3.3-8B calls unforced on 0.61, writing most of the rest as a code block, and Phi-4-mini on 0.70,
announcing the tool in prose on the rest, so both stay out.

\paragraph{The two tests, fixed before the tasks were built.} Among test tasks correct under both
full-precision instruments, invocation damage must exceed selection damage by at least 15 points on
at least three builds, and the gap averaged equally over the builds must have a 95\% lower bound
above 10 points when whole tasks are resampled; round-to-nearest and calibrated runs are reported
apart, and a claim about deployment needs the calibrated runs to repeat the result. For prediction, the law must beat the calibration tasks'
prevalence by at least 10\% in Brier score on emitted failures, with a paired interval excluding
zero. Both are read from the stored runs.

\begin{table}[h]
\caption{Invocation and selection losses on the same subset of the 320 sealed tasks: those
correct under both full-precision instruments. Each rate uses that shared denominator, which is
260 tasks on Hermes-3, 285 on Qwen3-1.7B and 300 on the other models; gaps use unrounded rates. A call counts when it is opened and names a tool, in the model's own format. The test fixed in advance asks for a gap of at least 15 points on three builds and a lower bound above 10 on the five-build average; the round-to-nearest runs give both. The two calibrated runs give neither.}
\centering\small
\begin{tabular}{llrrrr}
\toprule
model & 3-bit run & calls lost & names lost & gap (points) & 95\% lower \\
\midrule
Gemma-3-4B & round-to-nearest & 0.87 & 0.32 & $+55.3$ & $+48.7$ \\
Qwen2.5-7B & round-to-nearest & 0.57 & 0.12 & $+44.3$ & $+37.7$ \\
Qwen3-4B & round-to-nearest & 0.75 & 0.55 & $+19.7$ & $+12.5$ \\
Qwen3-1.7B & round-to-nearest & 0.93 & 0.96 & $-3.5$ & $-7.4$ \\
Hermes-3-Llama-3.2-3B & round-to-nearest & 0.44 & 0.59 & $-15.0$ & $-22.7$ \\
\midrule
all five, equally weighted & & & & $+20.2$ & $+17.1$ \\
\midrule
Qwen2.5-7B & AutoRound-GPTQ & 0.18 & 0.02 & $+15.7$ & $+11.0$ \\
Qwen3-4B & AutoRound-GPTQ & 0.00 & 0.01 & $-1.0$ & $-2.6$ \\
\bottomrule
\end{tabular}
\label{tab:crossed}
\end{table}

\paragraph{How the calls are lost.} The three builds that lose the call first lose it in different
ways, and the way matters for what a deployer sees. Gemma-3-4B and Qwen3-4B stop opening it, on
80\% and 57\% of the tasks they opened it on at full precision; Qwen2.5-7B opens it every time. Gemma-3-4B writes prose that names
the tool and never opens the call, ``Okay, I will use the \texttt{Hotels\_2\_SearchHouse}
$\ldots$'', which is a model declining to act. Qwen3-4B opens a call on 42\% of the test tasks against 97\% at full
precision, and finishes a parseable one on 24\%, so most of its loss is the decision and the
rest is the format. Qwen2.5-7B keeps opening the call on 97\% of them and
finishes a parseable name on 43\%, so its failure is the format breaking around a decision that
held. The calibrated Qwen2.5-7B build fails the same way one step milder: of its 53 lost calls, 41
emit the call token and then the end-of-turn token immediately, and its margins put the
whether-to-call sign change at 0.006. Both calibrated builds also call more often when the correct
tool has been removed, 0.31 against 0.16 on Qwen3-4B and 0.28 against 0.22 on Qwen2.5-7B, so
calibration moves the error from silence toward over-calling.

\paragraph{Margins on the same tasks.} On Qwen3-4B at 3 bits the two halves keep nearly the same
share, $c = 0.12$ for whether-to-call against $0.24$ for which-tool, and separate in their offsets,
$-14.6$ logits against $-0.6$. Gemma-3-4B repeats that shape at $c = 0.10$ against $0.19$ with
offsets $-3.6$ and $+0.5$. Qwen2.5-7B at 3 bits holds both, $c = 0.49$ and $0.51$, which is why its
lost calls are formation rather than decision. Hermes-3-Llama-3.2-3B is in the emitted results and not
here: at full precision only 0.44 of its should-invoke margins are positive on tasks it calls on
81\% of the time, so on the prompt its template forces the pair is close to a tie and reads
nothing.

\paragraph{Prediction.} Forecasting each test task's emitted invocation failure from its
should-invoke margin, with $(c,b,\sigma)$ fitted on the calibration tasks, beats the prevalence
forecast at 4 bits on Qwen3-4B by 37\% in Brier score (95\% CI 24 to 50) and fails everywhere else.
At 3 bits it fails in both directions: on Qwen2.5-7B the margins keep their sign and 57\% of calls
fail to form, and on Qwen3-4B nearly every margin changes sign while a quarter of the calls still
come out. The forecast test is therefore not met, and \S\ref{sec:forecast} states the forecast over margin
sign changes, which is what it is fitted and validated on.

\subsection{The forecast, in full}
\label{app:forecastfull}

\paragraph{Against the calibration half's own rate.} A deployer who has run the calibration half already holds that half's observed flip rate, so the forecast has to beat it to be worth fitting. Scored on the same held-out halves and the same 1{,}082 conditions as the held-out forecast of \S\ref{sec:forecast}, the law's median error is 1.0 percentage points against 1.3 for the fitting half's own rate, 7.8 against 9.0 at the 90th percentile, and the law is closer on 47\% of conditions with 20\% tied. On the 316 conditions whose observed rate lies between 0.2 and 0.8 the gap opens: 4.3 against 6.0 at the median, the law closer on 66\%, and an item-level Brier score of 0.186 against 0.239. Which-tool and argument filling repeat requests, so a split that cuts between repeats puts one request on both sides. Splitting instead by the request as posed, with the scored option pair excluded from its identity, keeps the law ahead of the fitting half's rate and widens the middle-range gap to 2.5 against 4.6. That grouping covers the 802 conditions whose items can be traced back to the request that produced them, and no request straddles it; the rule it replaced, which grouped items by their full-precision margin, leaked a request across 249 of those same conditions.

\begin{figure}[t]
\begin{center}
\includegraphics[width=\linewidth]{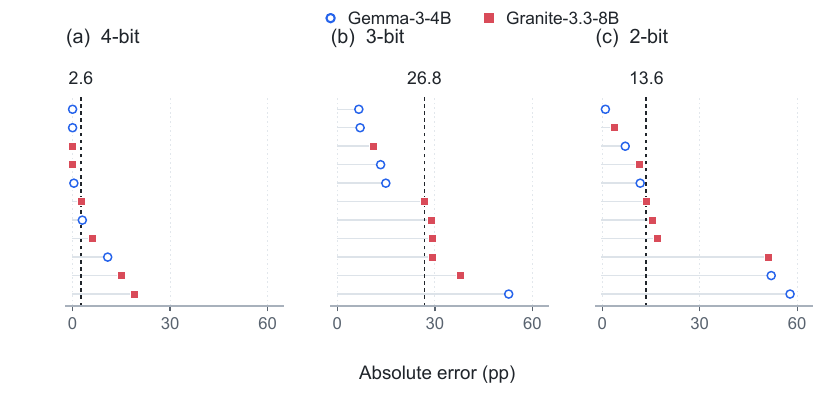}
\end{center}
\caption{Borrowed constants, applied to one model held out of the matrix and one weaker control:
Gemma-3-4B has no tool or safety decisions in the matrix; Granite-3.3-8B has both there. Eleven forecasts per bit-width are ranked by absolute error; marker shape identifies the model
and each labelled dashed line marks the median. At 4 bits the median error is 2.6
points and at 3 bits it is 26.8, where a deployer has to measure their own model rather than
borrow ours.}
\label{fig:transfer}
\end{figure}

\paragraph{The calibration pass, condition by condition.}
The limits appear when constants transfer across models, and we measured them. Applying another model's fitted constants to the two transfer-test models gives 1.7 to 8.4 percentage points of error at 4 bits and 9.5 to 33 at 3 bits, taking the range of mean errors across model-and-test-set pairs; one of the two (Gemma-3-4B) has no tool or safety decisions in the result matrix, while the other (Granite-3.3-8B) has both there and is the weaker test. The forecast is accurate where nothing happens and wrong where damage begins (Figure~\ref{fig:transfer}). One calibration pass per model repairs exactly the case that needs it: fitting $(c,b,\sigma)$ on half a condition's decisions and scoring the other half cuts the median error from 11.1 percentage points to 3.4 over 71 scoreable conditions on eight models, and the gain is not uniform. At 4 bits it helps in fewer than half the conditions, 16 of 36, and moves the median only from 4.0 points to 3.6, because constants borrowed from another model already work where quantization has broken nothing. At 3 bits it takes 14.4 points to 2.7 and helps in 16 of 18 conditions, and at 2 bits 16.9 to 1.9. On whether-to-call it helps in \emph{every} damaged condition, 7.1 to 2.1 at 3 bits and 24.6 to 1.1 at 2. Six of the eight models improve on their own median, Phi-4-mini stays level (3.7 to 3.8) and Gemma-4-E4B gets worse, 5.3 points to 8.5. Sixty-four of the 135 conditions cannot be scored because the slope is not identifiable on half the data. Borrowing across conditions rather than across models says the same thing on a much larger
set of fits. Holding out one whole condition at a time from the 1{,}580 estimable ones and
predicting its flip rate from the median constants of every other condition at its bit-width,
the median absolute error is 9.8 points at 3 bits against 1.3 for the condition's own
constants, and 12.6 when the donors come from every bit-width. At 8 bits the same borrowing costs 1.6 points, because
little has moved for the constants to be wrong about. So the constants are a property of the
condition at the bit-width where the damage is, which is the case for running the pass rather
than reading ours.

\paragraph{Forecast accuracy, in full.}
The median is flattered by conditions whose observed rate already sits at or near an endpoint, and the distribution behind it is heavy-tailed. The whole tail runs 3.5 points at the 75th percentile, 7.8 at the 90th and 10.4 at the 95th; by damage band the median error is 0.5 at $c \ge 0.70$, 1.3 in 0.30--0.70, 3.3 in 0--0.30 and 3.3 below zero; and the 766 conditions whose observed rate is at or near an endpoint carry a median of 0.5, which is what flatters the headline. Read as a per-item uncertainty estimate, the formula is well calibrated: pooling 193{,}779 item-level predictions from 1{,}430 estimable conditions, each from its condition's own fit, decisions given a flip probability of $p$ flip at a frequency within 0.021 of $p$ in every probability band (expected calibration error 0.005; Brier score 0.079 against 0.165 for guessing the average rate). Conditions from the same model are not independent. Appendix~\ref{app:accounting} reports the held-out forecast and the other summaries with each model counted once.

\subsection{Condition accounting}
\label{app:accounting}

The main \emph{matrix} contains 1{,}183 conditions. Its 86 runs cover mostly weight-quantization experiments. Additional experiments test activations, rotation and the KV cache. The model comparison uses 357 runs; the held-out forecasts and activation analyses also draw on additional runs. Table~\ref{tab:accounting} lists the population for the main analyses. A condition is one model configuration evaluated at one bit-width on one kind of decision in a single run. Repeated runs remain separate, and flip rates use the readable items in each group.

For the three kinds whose items have a should and a should-not side, the matrix holds each condition three times: both sides together and each side alone. Dropping the pooled entries leaves the medians of \S\ref{sec:attenuation} unchanged. Conditions from one model are not independent observations, so the law's in-sample error and the one-more-bit baseline are also reported clustered: each model's median taken first, then the median of those, which counts a model with many runs once. Checking Eq.~\ref{eq:flip} against each condition's own observed flip rate, in sample, the error moves from 0.75 points per condition over 1{,}409 conditions to 0.97 per model over 23 models, and the one-more-bit baseline from 0.330 over 458 conditions to 0.353 over 12 models. Both changes are small; both summaries are fitted and scored on the same items. The held-out forecast survives the same treatment: 1.4 points per model over 20 models against 1.0 per condition, on the 1{,}053 of its 1{,}082 conditions whose model is recoverable. The rival comparison needs no such check, since the multiplicative account wins every damaged condition. Four runs whose model cannot be recovered from their records are excluded rather than pooled into an unlabelled bucket.

\subsection{The verdict by way of quantizing and bit-width}
\label{app:strata}

\begin{table}[!ht]
\caption{Damaged core-kind conditions ($c<0.70$, slope estimable) against how far the margin is anchored to behaviour there. Anchoring is the full-vocabulary argmax share retained relative to the same model and test set at full precision, since the safety test set's absolute share is low (21--24\%) even undamaged. Rotated 4-bit activations keep the anchoring (79--80\%) and rotated 3-bit ones do not (0--15\%).}
\label{tab:strata}
\begin{center}
\footnotesize
\setlength{\tabcolsep}{4pt}
\begin{tabular}{@{}lrrl@{}}
\toprule
Condition & Conditions & M wins & Anchoring \\
\midrule
Weight $\ge$4 bits             & 35 & 35 & 94--97\% \\
Activation A6 tool, rotated A4 & 13 & 13 & 79--80\% \\
Activation A6, safety          &  9 &  9 & 53\% \\
Activation $\le$A5, rotated A3 & 93 & 93 & 0--15\% \\
Weight 3 and 2 bits            & 203 & 203 & 0--34\% \\
Unresolved (A8, rot.\ A5, KV)  & 14 & 14 & not run \\
\midrule
Total & 367 & 367 & \\
\bottomrule
\end{tabular}
\end{center}
\end{table}

The free-generation check that anchors the margin to behaviour was run on the weight axis only; the
full-vocabulary check was also run on the activation axis (Appendix~\ref{app:anchor}), and both verdicts depend on bit-width
(Appendix~\ref{sec:instrument}). If the model comparison of
\S\ref{sec:rivals} drew its result from the depths where those checks are weakest, the result would
be an artifact of the instrument rather than a finding about quantization. It does not
(Table~\ref{tab:strata}): the multiplicative account takes every band unanimously, including the
48 conditions in the two bands that keep 79\% or more of the full-precision argmax share: 35 at 4-bit-and-above weights, the only
setting where the free-generation check was also run, and 13 in the rescued activation group (9 at rotated 4-bit activations, 4 at naive 6-bit tool), whose band was measured on Qwen3-4B's tool set.

The stratification also shows where the margin may be read as behaviour (Appendix~\ref{app:anchor}). Activation quantization is therefore not simply a second axis of
the same damage: without the rotation it leaves the instrument's behavioural anchoring far earlier
than weight rounding does, and the 93 conditions at 5-bit activations and below, or at 3-bit activations with the rotation, carry the same caveat as the 3- and 2-bit weight results. The fourteen unresolved conditions are held out of these bands: 8-bit activations, rotated A5, and the one that quantizes the key--value cache rather than the layer inputs. Their anchoring has not been measured, and the cache condition is a different intervention from the ones the bands were measured on.

\end{document}